\documentclass[10pt,journal,compsoc]{IEEEtran}

\ifCLASSOPTIONcompsoc
\usepackage[nocompress]{cite}
\else
\usepackage{cite}
\fi
\ifCLASSINFOpdf
\usepackage[pdftex]{graphicx}
\else
\fi
\usepackage{amsmath}
\usepackage{amsthm}
\usepackage{amssymb}
\usepackage{algorithm}
\usepackage{algorithmic}
\usepackage{multirow}
\usepackage[table]{xcolor}

\usepackage{url}

\begin{document}
	\newcommand{\bfx}{{\mathbf{x}}}
	\newcommand{\bfh}{{\mathbf{h}}}
	\newcommand{\bfw}{{\mathbf{w}}}
	\newcommand{\bfg}{{\mathbf{g}}}
	
	\newcommand{\bfX}{{\mathbf{X}}}
	
	\newcommand{\calC}{{\mathcal{C}}}
	\newcommand{\calO}{{\mathcal{O}}}
	\newcommand{\calM}{{\mathcal{M}}}
	
	\newcommand{\calR}{{\mathcal{R}}}
	\newcommand{\calL}{{\mathcal{L}}}
	\newcommand{\calI}{{\mathcal{I}}}
	
	\newcommand{\calK}{{\mathcal{K}}}
	\newcommand{\calD}{{\mathcal{D}}}
	\newcommand{\calP}{{\mathcal{P}}}
	\newcommand{\calS}{{\mathcal{S}}}
	
	\newtheorem{property}{Property}
	\newtheorem{propsition}{Propsition}
	\newtheorem{theorem}{Theorem}
	\newtheorem{definition}[theorem]{Definition}
	
	\title{MAP: Model Aggregation and Personalization in Federated Learning with Incomplete Classes}

	\author{
		Xin-Chun~Li,
		Shaoming~Song,
		Yinchuan~Li,~\IEEEmembership{Member,~IEEE,}
		Bingshuai~Li,
		Yunfeng~Shao,
		Yang~Yang,
		and~De-Chuan~Zhan 
		\IEEEcompsocitemizethanks{
			\IEEEcompsocthanksitem 
			Xin-Chun Li and De-Chuan Zhan are with the State Key Laboratory for Novel Software Technology, Nanjing University, Nanjing, Jiangsu 210023, China.
			E-mail: lixc@lamda.nju.edu.cn, zhandc@nju.edu.cn
			\IEEEcompsocthanksitem 
			Yang Yang is with the Nanjing University of Science and Technology, Nanjing, Jiangsu 210094, China.
			E-mail: yyang@njust.edu.cn
			\IEEEcompsocthanksitem 
			Shaoming Song, Yinchuan Li, Bingshuai Li, and Yunfeng Shao are with Huawei Noah’s Ark Lab, Beijing 100085, China.
			E-mail: \{shaoming.song, liyinchuan, libingshuai, shaoyunfeng\}@huawei.com
		}
		\thanks{Yang Yang is the corresponding author.}
	}

	\IEEEtitleabstractindextext{%
		\begin{abstract}
			In some real-world applications, data samples are usually distributed on local devices, where federated learning (FL) techniques are proposed to coordinate decentralized clients without directly sharing users' private data. FL commonly follows the parameter server architecture and contains multiple personalization and aggregation procedures. The natural data heterogeneity across clients, i.e., Non-I.I.D. data, challenges both the aggregation and personalization goals in FL. In this paper, we focus on a special kind of Non-I.I.D. scene where clients own incomplete classes, i.e., each client can only access a partial set of the whole class set. The server aims to aggregate a complete classification model that could generalize to all classes, while the clients are inclined to improve the performance of distinguishing their observed classes. For better model aggregation, we point out that the standard softmax will encounter several problems caused by missing classes and propose ``restricted softmax" as an alternative. For better model personalization, we point out that the hard-won personalized models are not well exploited and propose ``inherited private model" to store the personalization experience. Our proposed algorithm named MAP could simultaneously achieve the aggregation and personalization goals in FL. Abundant experimental studies verify the superiorities of our algorithm.
		\end{abstract}

		}

	\maketitle

	\IEEEdisplaynontitleabstractindextext

	%
	\IEEEpeerreviewmaketitle

	\IEEEraisesectionheading{\section{Introduction}\label{sec:introduction}}
	
	\IEEEPARstart{A}{lthough} deep learning has experienced great success in many fields~\cite{imagenet,ResNet}, a data center training paradigm is usually required. Due to data privacy or transmission cost, data from individual participants can not be located on the same device in some real-world applications~\cite{Gboard}. Standard distributed optimization~\cite{ParameterServer,LargeScaleNet} provides solutions to distributed training with big data or huge models, while federated learning (FL)~\cite{Fed-Concept,FedAvg,zhang2022personalized,FedPS,li2022mining,li2021personalized,FLAGG-TKDE} is tailored for data privacy protection and efficient distributed training. Like most distributed optimization algorithms, FL is commonly organized in the parameter server architecture~\cite{ParameterServer} where a server coordinates amounts of clients to accomplish the model training process. Specifically, FL takes multiple rounds of local personalization and global aggregation procedures~\cite{FedAvg} to coordinate isolated clients. During the local personalization procedure, a subset of clients download the global model from the central server and update it on their local private data. During the global aggregation procedure, the central server receives clients' model parameters and updates the global model. These two procedures are iterated until convergence. In the whole process, only model parameters are transmitted among clients and the server, which brings basic privacy protection for participating users. Additionally, clients often undertake more computation steps during local personalization procedures, e.g., epochs of training on local data, making decentralized training more communication efficient. Due to these advantages, FL has been successfully applied to computer vision~\cite{FedCV,FedPAN}, natural language processing~\cite{FedNLP,FedSent}, speech analysis~\cite{FedKWS-UI}, etc.
	
	
	The server and clients in FL have different goals. The server aims to aggregate a single global model that could generalize well to forthcoming data, while clients participate in FL with the purpose of enhancing the local model performance on their own private data. To be brief, the server wants to obtain a single global model that performs well on the global data distribution, while clients want to obtain personalized models that perform well on their local data distributions. The Non-I.I.D. data across clients~\cite{FedAvg,FedProx,Fed-Concept} hinders both the global aggregation and local personalization goals in FL. On the one hand, the local training procedure will diverge a lot from the global target due to the discrepancy between the local and global data distribution, leading to the weight divergence phenomenon that degrades the aggregation performance~\cite{Fed-NonIID-Data}. On the other hand, the globally aggregated model may be a compromise among clients, which performs worse than the locally trained model on a specific client's data and dispels the client's incentive to participate in FL~\cite{FedAdapt}. The stronger the heterogeneity of the local data distributions, the harder it is to simultaneously obtain a satisfied global model and local personalized models. Hence, it is necessary to design effective strategies for both aggregation and personalization goals under Non-I.I.D. scenes.
	
\begin{figure}[tbp]
		\centering
		\includegraphics[width=\linewidth]{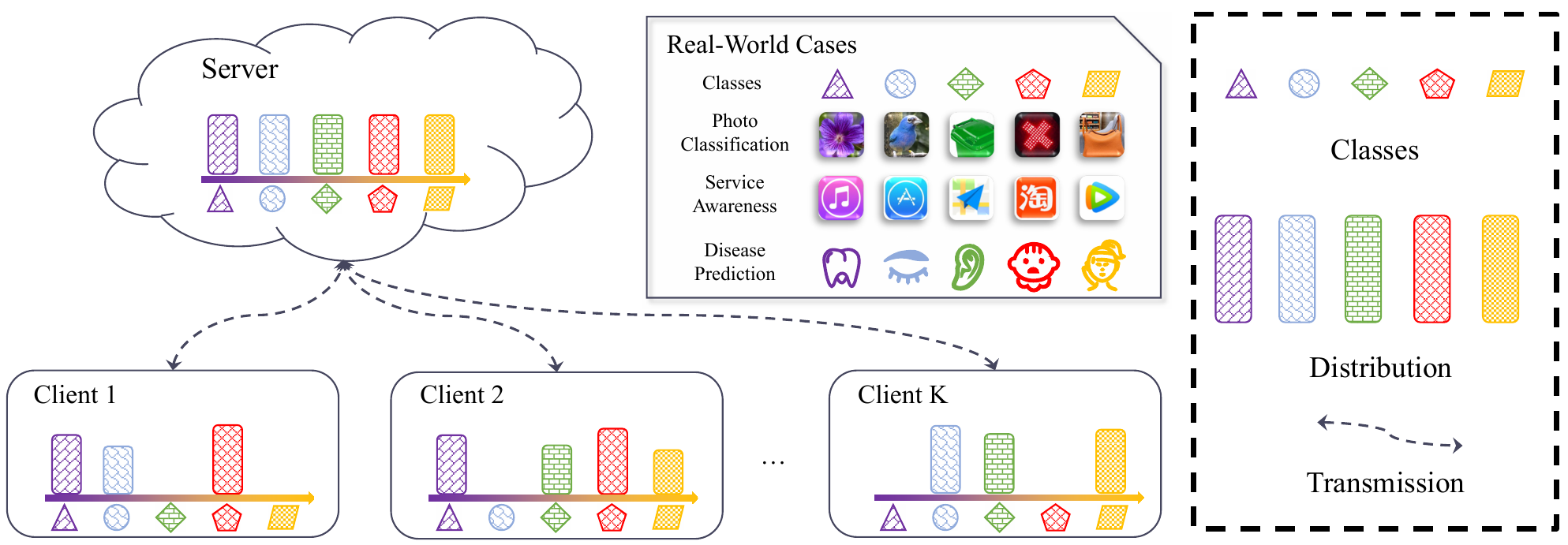}
		\caption{{\small The considered FL scene with incomplete classes.}}
		\label{fig:scene}
	\end{figure}

	In this paper, we study a special kind of Non-I.I.D. scene where clients own incomplete classes. Specifically, the server is targeted to classify a set of classes, while clients only observe a partial set of these classes. This scene is an extreme case of the label distribution skew~\cite{BNFL,POSFL,AddressImb,FedDF} in FL that certain classes are not presented in some clients' data currently or even in the future. Hence, the server's goal is to obtain a model that could distinguish all classes well, while clients only aim to obtain personalized models that perform well on their observed classes. For example, in photo classification, the server needs to recognize many types of photos, while a specific terminal device only needs to recognize partial classes~\cite{FedProto}; in service awareness application, the server needs to build a model to identify the byte stream of multiple APP types, while a user may not have downloaded some APPs and does not need to identify them~\cite{FedRS}; in disease prediction, the server needs to obtain a model to recognize disease types as more as possible, while a specialized hospital only needs to recognize a fraction of disease types. The illustrations of the scene and goals are shown in Fig.~\ref{fig:scene}. Our motivation is to achieve both aggregation and personalization goals in the FL scene with incomplete classes. Fortunately, the author's two previous works, i.e., FedRS~\cite{FedRS} and FedPHP~\cite{FedPHP}, can be seamlessly combined to achieve both goals.
	
	FedRS~\cite{FedRS} is proposed for better model aggregation, which in-depth analyzes and improves the classification layer of a neural network. Existing studies show that the topmost layer is more task-specific than bottom layers~\cite{NN-Transferability}. Faced with label distribution skew, the commonly utilized softmax operation for classification~\cite{Softmax-MultiStage} could be more vulnerable. Specifically, we first analyze and show the inherent pulling and pushing forces in standard softmax. Then we point out that several properties will disappear faced with missing classes, making the missing classes' proxies become inaccurate during local training procedures of FL. As an alternative, we slightly modify the standard softmax and introduce ``restricted softmax" (RS) to limit the update of missing classes' weights for better aggregation. 
	
	FedPHP~\cite{FedPHP} is proposed for better model personalization, which in-depth analyzes and improves the exploitation of personalized models. Our significant observation is that the newly downloaded global model from the server may perform poorly on clients, while it could become better after adequate personalization steps. Inspired by this, we advocate that the hard-won personalized model in each communication round should be rationally exploited, while standard FL methods directly overwrite the previous personalized models. Specifically, we propose a novel concept named ``inherited private model" (HPM) for each local client as a temporal ensembling of its historical personalized models and exploit it to supervise the personalization process in the next global round.
	
	FedRS is designed for model aggregation, while FedPHP is for model personalization, and neither of them could achieve aggregation and personalization goals simultaneously. Hence, we utilize the ``RS" in FedRS for better model aggregation and the ``HPM" in FedPHP for better model personalization, which could simultaneously achieve model aggregation and personalization goals in FL with incomplete classes. We name the proposed method MAP and investigate its superiorities via abundant experimental studies and analysis.
	
	\section{Related Works}
	\noindent \textbf{FL with Non-I.I.D. Data.}
	Commonly, FL divides the Non-I.I.D. scenes into five categories: feature distribution skew, label distribution skew, concept shift with different features, concept shift with different labels, and quantity skew. The first two categories correspond to the covariate shift and label shift in dataset shift~\cite{DatasetShift}. Label distribution skew is also related to class imbalanced learning with long tail data~\cite{Imbalance-DA,Imbalance-Margin,AddressImb,AgnosticFL-Imbalance} or class incremental learning with new classes~\cite{FedINC,KDD-Inc,TKDE-Inc}. Label distribution skew in FL refers to that the prior label distributions $\calP^k(y)$ may vary a lot across clients, but $\calP^k(\bfx | y)$ is the same. The considered scene in this paper is a special case of label distribution skew in FL, where the prior label distribution of some clients is extremely imbalanced, i.e., some classes are missing. Various techniques have been proposed to solve the Non-I.I.D. challenge in FL. Most of these techniques improve the aggregation and personalization performances independently. One exception is the recently proposed FedROD~\cite{FedROD}, which borrows techniques from class-imbalanced learning and private-shared models for both aggregation and personalization in FL. Different from FedROD, we in-depth analyze the drawbacks of traditional FL algorithms and propose novel techniques.
	
	\noindent \textbf{Model Aggregation with Non-I.I.D. Data.}
	FedAvg~\cite{FedAvg} takes the simplest parameter averaging for model aggregation, whose performance may degrade under Non-I.I.D. data. The heterogeneity of clients' data distributions could make the local updates too diverged to aggregate, i.e., the phenomenon of weight divergence~\cite{Fed-NonIID-Data}. Sharing a small public data set among clients could benefit model fusion with a slight risk of privacy leakage~\cite{Fed-Shared-Data,Fed-KeepTrace,HMR,FedDF}. Utilizing local regularizations~\cite{FedProx,FedDyn,FedMMD} or control variates~\cite{Scaffold} could restrict the local model parameters not diverge from the global model too much. Designing better optimization strategies could also help aggregation~\cite{FedOpt,FedNova}. Some methods aim to align model parameters for better aggregation~\cite{BNFL,FedMA,FedPAN}. FedAwS~\cite{FedAwS} studies the extreme scene where each client could only access one class. The author's previous work, FedRS~\cite{FedRS}, proposes ``RS" to handle the missing classes for better model aggregation. Utilizing class-imbalanced learning techniques is also a possible solution for better aggregation~\cite{FedROD,FedRealCV}.
	
	\noindent \textbf{Model Personalization with Non-I.I.D. Data.}
	FedAvg~\cite{FedAvg} generates a single global model, which is hard to capture heterogeneous local distributions simultaneously~\cite{FedBoost}. Finetuning the aggregated models is the most direct solution for personalization~\cite{EvalPer,FedAdapt}. Some methods resort to multi-task learning~\cite{Fed-MultiTask} or meta learning~\cite{PerFedAvg,Fed-MAML} for fast local adaptation. Taking advantage of fully decentralized learning~\cite{Fed-Decentralize} or private-shared models~\cite{FLDA,FedRep,L2GD,FedROD} are also solutions for better personalization. The author's previous work, FedPHP~\cite{FedPHP}, proposes ``HPM" to store the historical personalization experience for better model personalization. A previous work~\cite{FedPS} empirically studies how to design appropriate shared-private network architectures for aggregation and personalization. The recent work~\cite{TowardsPFL} presents a unique taxonomy of personalized federated learning techniques.
	
	\section{Background and Issues to Solve}
	\subsection{Basic Procedure of FL} \label{sect:fl-procedure}
	In standard FL systems, e.g., FedAvg~\cite{FedAvg}, we have $K$ distributed clients and a global parameter server. Each client could be an organization, an individual user, or a terminal device, etc. The $k$-th client has a local data distribution $\calD^k=\calP^k(\bfx, y)$, and $\calD^k$ could differ from other clients' distributions a lot, where $k \sim \calK = \{1,2,\cdots,K\}$. For the $k$-th client, its optimization target is:
	\begin{equation}
		\calL(\psi^k) = \mathbb{E}_{(\bfx^k, y^k) \sim \calD^k}\left[ \ell\left( f(\bfx^k;\psi^k), y^k \right) \right], \label{eq:local-target}
	\end{equation}
	where $f(\cdot;\psi^k)$ is the prediction function with parameters $\psi^k$. $\ell\left(\cdot, \cdot\right)$ is the loss function, which is usually the cross-entropy loss for classification tasks. FedAvg takes the following optimization target:
	\begin{equation}
		\calL(\psi) = \sum_{k=1}^K p_k \left[ \mathbb{E}_{(\bfx^k, y^k) \sim \calD^k}\left[ \ell\left( f(\bfx^k;\psi), y^k \right) \right] \right], \label{eq:global-target}
	\end{equation}
	where $p_k$ denotes the weight of the $k$-th client, which satisfies $\sum_{k=1}^K p_k = 1$. We denote $\psi$ as the global parameters on the server, and $\psi^k$ as the parameters on the $k$-th client. The optimization target in Eq.~\ref{eq:global-target} can be solved by rounds of local personalization and global aggregation procedures. During the local procedure in $t$-th round, where $t \in \{1, 2, \cdots, T\}$, a subset of clients $S_t$ is selected, and the selected clients download the global model, i.e., $\psi_t^k \leftarrow \psi_{t}$. Then, the $k$-th client samples data batches from its training data and updates $\psi_t^k$ via the corresponding empirical loss of Eq.~\ref{eq:local-target}. The updated parameters, denoted as $\hat{\psi}_t^k$, will be uploaded to the server. During the aggregation procedure, the server averages the global model via $\psi_{t+1} \leftarrow \sum_{k \in S_t} \frac{1}{|S_t|} \hat{\psi}_{t}^k$. These two procedures will repeat $T$ communication rounds until the convergence.
	
	\subsection{FL Scene with Incomplete Classes} \label{sect:fl-scene}
	The considered FL scene with incomplete classes is an extreme case of label disribution skew in FL. Suppose the $k$-th client owns a training set $D^k_{\text{tr}}=\{(\bfx^k_i, y^k_i ) \}_{i=1}^{N^{k}_{\text{tr}}} \sim \calD^k$ with $N^{k}_{\text{tr}}$ samples, and a test set $D^k_{\text{te}}=\{(\bfx^k_j, y^k_j ) \}_{j=1}^{N^{k}_{\text{te}}} \sim \calD^k$ with $N^{k}_{\text{te}}$ samples. According to the assumption of label distribution skew in FL, only the prior class distribution $\calP^k(y)$ varies across clients, while $\calP^k(\bfx | y)$ keeps the same. To be an extreme case, we assume that $\calP^k(y=c) = 0$ holds for some certain classes $c \in \calC = \{1, 2, \ldots, C\}$. $C$ is the number of classes. That is, the $k$-th client's data could only come from an incomplete class set, i.e., $y^k_i$ and $y^k_j$ is only from $\calO^k \subseteq \calC$. $\calO^k$ denotes the observed class set of the $k$-th client. Similarly, the missing class set is denoted as $\calM^k = \calC \setminus \calO^k$. Note that the $k$-th client actually does a $|\calO^k|$-class classification problem. We also assume $\cup_{k=1}^K \calO^k = \calC$, which means that the server aims to obtain a complete classification model that distinguishes samples from $\calC$, i.e., a $C$-class classification model. Hence, to evaluate the aggregated global model, the server owns a global test set denoted as $D_{\text{te}}=\{(\bfx_i, y_i ) \}_{i=1}^{N_{\text{te}}}$ with $N_{\text{te}}$ samples, where $y_i \in \calC$ and follows a uniform distribution. The scene is illustrated in Fig.~\ref{fig:scene}.

	\subsection{Goals of Aggregation and Personalization} \label{sect:fl-goals}
	The server aims to obtain a single global model that minimizes Eq.~\ref{eq:global-target}, i.e., a $C$-class classification model. The $k$-th client is inclined to obtain a well-performed model that could minimize the Eq.~\ref{eq:local-target} on its own data distribution, i.e., a $|\calO^k|$-class classification model. We formally present the goals and corresponding metrics of aggregation and personalization as follows:
	\begin{definition}[Goal of Aggregation in FL] \label{def-aggregation}
		The parameter server calculates the classification accuracy of the aggregated model on the global test set $D_{\text{te}}$ in each communication round to evaluate the aggregation performance.
	\end{definition}
	
	\begin{definition}[Goal of Personalization in FL] \label{def-personalization}
		The $k$-th client calculates the performance of the personalized model on the local test set $D^k_{\text{te}}$ in each communication round. The average accuracy across selected clients evaluates the personalization performance.
	\end{definition}
	
	During each communication round, we distinguish the model before and after personalization as the newly-downloaded global model and the personalized model, respectively. Notably, we calculate the personalization performance with the personalized model rather than the newly-downloaded model.

	We can find that the goal of aggregation emphasizes the globally aggregated model's ability, which is expected to work well universally for existing clients or novel clients. However, the goal of personalization emphasizes the personalized model's ability, which reflects the original incentive for clients to participate in FL. Due to the Non-I.I.D. challenge, it is hard to generate a well-performed aggregated model~\cite{Fed-NonIID-Data}. Even we can obtain such a global model, it could still perform poorly on clients, and delicate personalization methods should be further utilized for better adaptation~\cite{FedBoost,ThreePer}. Hence, we should design an effective FL method that could simultaneously improve both aggregation and personalization performances.

	\section{Proposed Methods}
	This section will introduce the proposed ``RS" for model aggregation, the ``HPM" for model personalization, and the proposed method MAP.
	
	\subsection{Restricted Softmax for Model Aggregation} \label{sect:rs-agg}
	
	\subsubsection{Softmax Properties} \label{sect:sf-property}
	Assume we have a training set $\{ ( \bfx_i, y_i ) \}_{i=1}^N$ with $N$ samples, where $y_i \in \calC = \{1, 2, \ldots, C\}$. Deep networks contain the feature extractor $F_{\theta}( \cdot )$ with parameters $\theta$ and the last classification layer with weights $\{\bfw_{c}\}_{c=1}^C$ (we omit the bias for simplification). The whole parameter set is denoted as $\psi = \left(\theta, \{\bfw_{c}\}_{c=1}^C\right)$. Without additional declaration, we refer to the last classification layer as the classifier. We denote $\bfh_i = F_{\theta}( \bfx_i ) \in \calR^{d}$ as the extracted feature vector of the $i$-th sample. We refer to the classification weights $\{\bfw_{c}\}_{c=1}^C$ as proxies. For the $c$-th proxy $\bfw_c$, we denote the features from the $c$-th class and other classes as positive features and negative features respectively.
	
	The commonly utilized softmax operation and cross-entropy loss are:
	\begin{eqnarray}
		p_{i,c} &=& \exp( \bfw_{c}^T \bfh_i )/\sum_{j=1}^C \exp( \bfw_{j}^T \bfh_i ), \label{eq:softmax} \\
		\calL &=& - \sum_{i=1}^N \sum_{c=1}^C \calI\{y_i = c\} \log p_{i, c}, \label{eq:ce-loss}
	\end{eqnarray}
	where $\calI\{\cdot\}$ is the indication function. The update of $\bfw_c$ via gradient descent could be divided into pulling and pushing forces:
	\begin{equation}
		\bfw_c = \bfw_c + \underbrace{\eta \sum_{i=1, y_i=c}^N \left( 1 - p_{i,c} \right) \bfh_i}_{\text{pulling}} \underbrace{- \eta \sum_{i=1,y_i\neq c}^N p_{i,c} \bfh_i}_{\text{pushing}}, \label{eq:update-w}
	\end{equation}
	where $\eta$ is the learning rate. Hence, we can obtain the following properties:
	\begin{property}[Properties of Softmax] \label{prop:softmax}
		Classification with softmax has the following properties: (1) pulling proxies closer to positive features; (2) pushing proxies away from negative features.
	\end{property}
	
\begin{figure}[tbp]
		\centering
		\includegraphics[width=\linewidth]{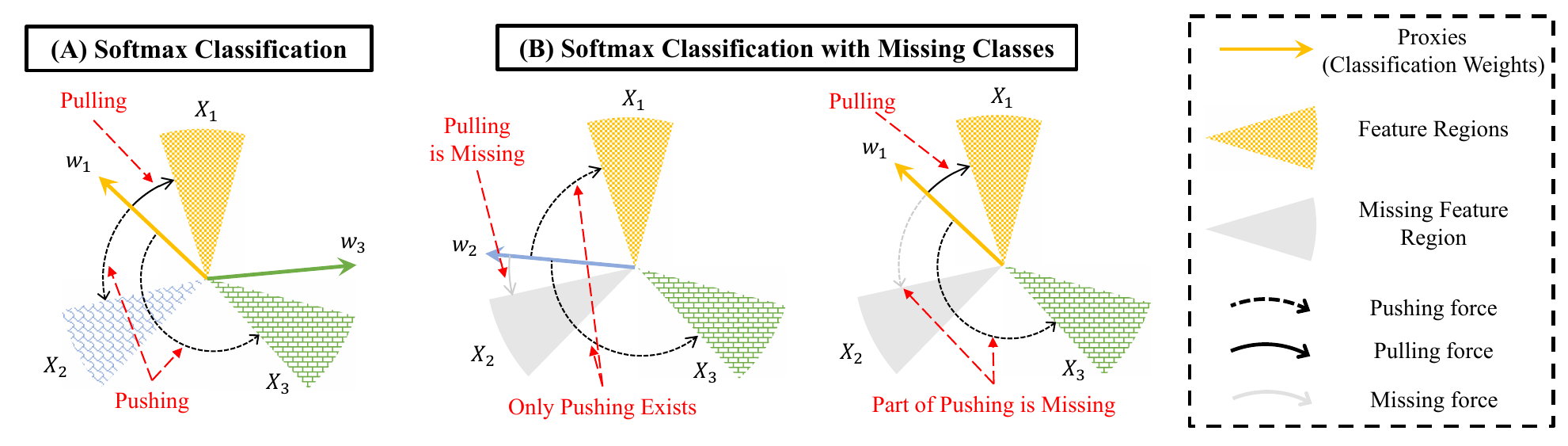}
		\caption{{\small The properties of softmax, i.e., Eq.~\ref{eq:update-w}, Eq.~\ref{eq:update-miss-w} and Eq.~\ref{eq:update-obser-w}.}}
		\label{fig:miss-sf}
	\end{figure}

	The properties are illustrated in Fig.~\ref{fig:miss-sf} (A), where a demo with three classes is shown. We only show the properties of updating proxy $\bfw_1$. The data region $\bfX_1$ contains positive features, while $\bfX_2$ and $\bfX_3$ contain negative features. Hence, $\bfw_1$ is pulled closer to $\bfX_1$ and pushed away from $\bfX_2$, $\bfX_3$ simultaneously. 
	
	\subsubsection{Softmax with Missing Classes} \label{sect:miss-sf-property}
	The above shows the softmax classification with complete classes. In the considered FL scene, the clients only observe a partial set of classes. We in-depth analyze the local update of the proxies with missing classes. We still denote the training set as $\{ ( \bfx_i, y_i ) \}_{i=1}^N$, while $y_i$ is only from $\calO \subseteq \calC$. The missing class set is denoted as $\calM=\calC \setminus \calO$. Then, the Eq.~\ref{eq:update-w} can be adapted. For $c \in \calM$:
	\begin{equation}
		\bfw_c = \bfw_c + \underbrace{\eta \sum_{i=1, y_i=c}^N \left( 1 - p_{i,c} \right) \bfh_i}_{=0} \underbrace{-\eta \sum_{i=1,y_i\neq c}^N p_{i,c} \bfh_i}_{\text{only pushing exists}}. \label{eq:update-miss-w}
	\end{equation}
	
	This shows that due to lacking corresponding training samples, the pulling force of proxy $\bfw_c, c \in \calM$ is missing, and the proxy is only pushed away from negative features. This phenomenon is illustrated in the left part of Fig.~\ref{fig:teaser} (B), where $\bfw_2$ is the proxy of the missing class and the gray arc implies the pulling property is missing.

	For the observed class $c \in \calO$, we have the update of $\bfw_c$:
	\begin{equation}
		\bfw_c = \bfw_c + \underbrace{\eta \sum_{i=1, y_i=c}^N \left( 1 - p_{i,c} \right) \bfh_i}_{\text{pulling}} \underbrace{-\eta \sum_{i=1,y_i \neq c}^N p_{i,c} \bfh_i}_{\text{$\left(|\calO|-1\right)$ classes}}, \label{eq:update-obser-w}
	\end{equation}
	where we can find that the pushing force becomes weaker due to that the negative features only come from $|\calO|-1$ classes. This is illustrated in the right part of Fig.~\ref{fig:teaser} (B), where the negative feature region $\bfX_2$ does not exist and only $\bfX_3$ exists when updating the proxy $\bfw_1$.
	
	Obviously, this update process could diverge from the oracle proxies a lot, and it could become less and less accurate with more training steps. When diverged from the oracle ones, the obtained proxies will be harder to aggregate. We will detail on this in the following sections. We conclude the above analysis as a problem of softmax with missing classes:
	\begin{property}[Problem of Softmax with Missing Classes] \label{prop:softmax-miss-class}
		With missing class set $\calM$, the softmax classification has following problems: (1) the proxies of missing classes, i.e., $\{\bfw_c\}_{c \in \calM}$, are only pushed away from negative features, becoming more and more inaccurate; (2) the proxies of observed classes, i.e., $\{\bfw_c\}_{c \in \calO}$, are only pushed away from $|\calO|-1$ negative feature regions.
	\end{property}
	
	\subsubsection{Restricted Softmax} \label{sect:rs}
\begin{figure}[!t]
		\centering
		\includegraphics[width=\linewidth]{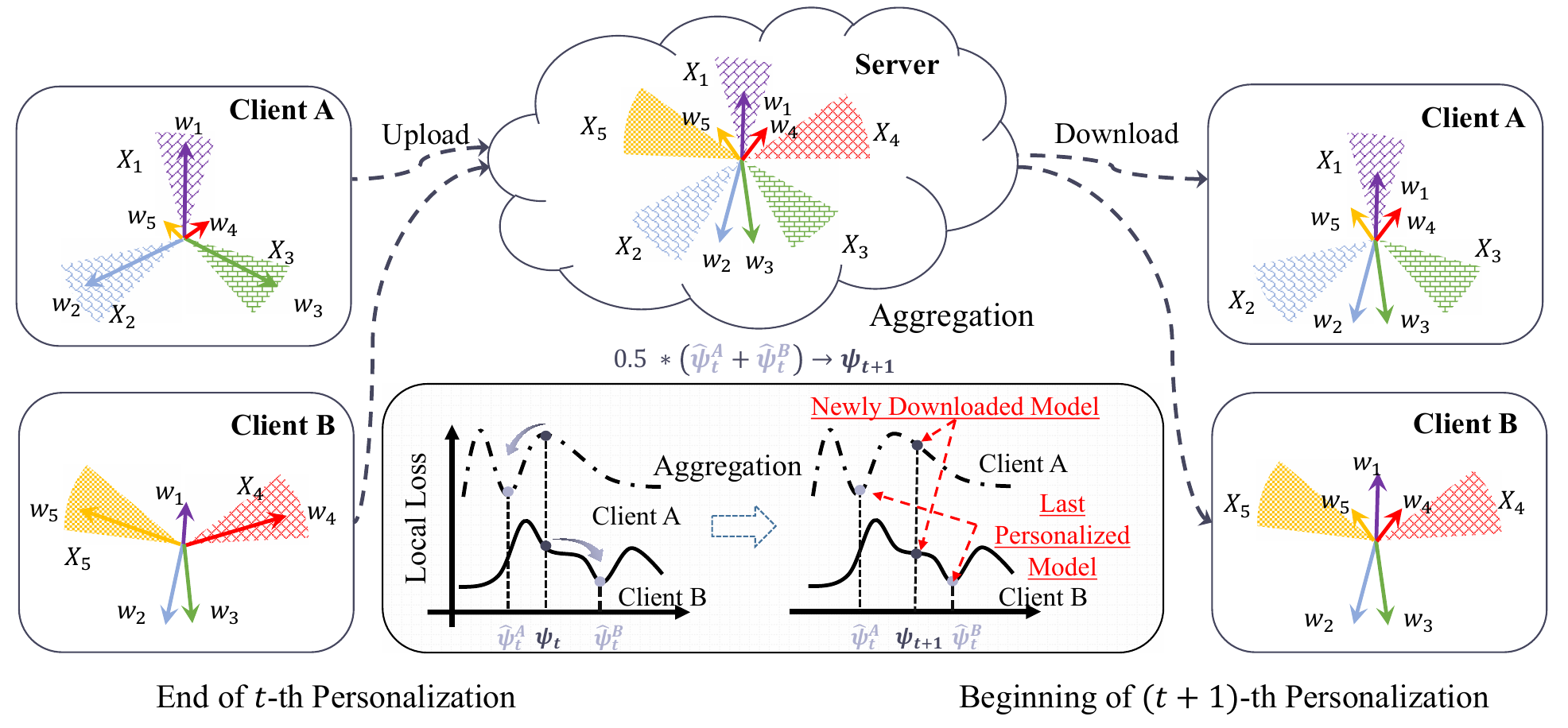}
		\caption{{\small Model aggregation and personalization challenges caused by incomplete classes.}}
		\label{fig:teaser}
	\end{figure}

	As introduced in Sect.~\ref{sect:fl-procedure} and Sect.~\ref{sect:fl-scene}, the selected clients will update the downloaded global model for multiple local epochs. The downloaded global model contains the full set of proxies $\{\bfw_c\}_{c=1}^C$, while the $k$-th client only observes a partial set (i.e., $\calO^k$) of the whole class set. This local training procedure with missing classes could cause several problems listed in Property~\ref{prop:softmax-miss-class}. That is, the missing classes' proxies will become inaccurate, making the models hard to aggregate. As illustrated in the upload stage in Fig.~\ref{fig:teaser}, the server aims to build a 5-class classification model while the two clients only observe 3 and 2 classes respectively. After the $t$-th personalization procedure, proxies $\bfw_4$, $\bfw_5$ on client A, and $\bfw_1$, $\bfw_2$, $\bfw_3$ on client B could become inaccurate, leading to a poor aggregation on the server, e.g., proxies $\bfw_2$, $\bfw_3$, $\bfw_4$, $\bfw_5$ diverge from their feature regions a lot. FL algorithms usually need amounts of communication rounds to converge. Hence, the error accumulation will become more serious, leading to an unstable training process and poor aggregation performances.
	
	To solve the problems caused by softmax, we advocate that the update of missing classes' proxies, i.e., $\{ \bfw_{c}^k \}_{c \in \calM^k}$, should be restricted. An easy way to implement this is adding ``scaling factors" to softmax operation, i.e.,
	\begin{equation}
		p^k_{i,c} = \exp( \alpha^k_c {\bfw^k_{c}}^T \bfh^k_i )/\sum_{j=1}^C \exp( \alpha^k_j {\bfw^k_{j}}^T \bfh^k_i ), \label{eq:rs}
	\end{equation}
	which is denoted as ``restricted softmax" (RS). We set $\alpha_{c}^k = \calI\{c \in \calO^k\} + \alpha \calI\{c \in \calM^k\}$, where $\alpha \in [0,1]$ is the only hyper-parameter. This is an asymmetric scaling way that works normally with $\alpha_{c}^k=1$ for observed classes while working as a decaying method with $\alpha_{c}^k=\alpha$ for missing classes. Although it is a simple modification, we in-depth analyze the brought advantages from several aspects. 
	
	\noindent \textbf{Restricting update of missing classes' proxies.} 
	Similar to Eq.~\ref{eq:update-miss-w}, we can obtain the gradient update of $\bfw^k_{c}$ under ``RS":
	\begin{equation}
		\bfw^k_c = \bfw^k_c \underbrace{-\alpha \eta \sum_{i=1}^{N_k} p^k_{i,c} \bfh^k_i}_{\text{restricted}}, \label{eq:update-miss-w-rs}
	\end{equation}
	where we can find that the pushing force is restricted with $\alpha \in [0, 1]$. If we take $\alpha = 0$, it degenerates into fixed proxies (if we do not consider weight decay); if $\alpha = 1$, it is just normal softmax. Overall, the update speed of missing classes' gradients is restricted. We use the ``RS" during the local training procedure and aggregate the model as common FL algorithms, e.g., the simple parameter averaging in FedAvg.
	
	\subsection{Inherited Private Model for Model Personalization} \label{sect:hpm-per}
	\subsubsection{Local Performance Degradation} \label{sect:per-obser}
	As the most standard FL algorithm, FedAvg~\cite{FedAvg} aims to generate a single global model, which is hard to capture heterogeneous local distributions simultaneously. An empirical observation in FedAvg is that the newly downloaded global model could perform poorly on local data. As shown in Fig.~\ref{fig:teaser}, at the beginning of the $(t+1)$-th personalization procedure, clients A and B first download the globally aggregated model (i.e., five proxies). Obviously, the downloaded proxies do not match the feature regions well, e.g., $\bfw_2$ and $\bfw_3$ deviate from the regions $\bfX_2$ and $\bfX_3$ significantly on client A. The newly-downloaded proxies could perform worse than the personalized models during the $t$-th personalization procedure. The loss curves in Fig.~\ref{fig:teaser} also disclose this phenomenon. Specifically, during the personalization stage of the $t$-th communication round, two clients first finetune the global model $\psi_t$ according to their own data distributions and obtain personalized models $\hat{\psi}_{t}^{\text{A}}$, $\hat{\psi}_{t}^{\text{B}}$ respectively. During aggregation, the server collects the personalized models and takes a direct parameter averaging as $\psi_{t+1} \leftarrow ( \hat{\psi}_{t}^{\text{A}} + \hat{\psi}_{t}^{\text{B}} )/2$. At the beginning of the next communication round, the aggregated model $\psi_{t+1}$ may perform worse than the last personalized models correspondingly, i.e., the local losses of $\psi_{t+1}$ are higher than $\hat{\psi}_{t}^{\text{A}}$ and $\hat{\psi}_{t}^{\text{B}}$. This phenomenon is named local performance degradation, and it happens once receiving the newly downloaded model from the server. A fundamental problem here is that the hard-won personalized models are directly overwritten by the newly downloaded global model (i.e., $\hat{\psi}_{t}^{\text{A}}$, $\hat{\psi}_{t}^{\text{B}}$ are not stored for better exploitation), and the clients have to personalize the global model from scratch in a new round. 
	
	\subsubsection{Inherited Private Model}
	To solve the local performance degradation problem, we propose a novel concept named ``inherited private model" (HPM) to keep the moving average of historical personalized models in each client and utilize it to supervise the newly downloaded model in the next round. Specifically, we use $\psi_{\text{p}}^k$ to denote the ``HPM" of the $k$-th client. At the beginning of the $t$-th personalization procedure, the $k$-th client downloads the global model: $\psi_{t}^k \leftarrow \psi_{t}$ and obtain the personalized model $\hat{\psi}_{t}^k$. The specific personalization process will be introduced later. At the end, we update $\psi_{\text{p}}^k$ via:
	\begin{equation}
		\psi_{\text{p},t+1}^k \leftarrow (1-\mu_{t}^k) \hat{\psi}_{t}^k + \mu_{t}^k \psi_{\text{p},t}^k, \label{eq:move-average}
	\end{equation}
	which keeps a moving average of historically personalized models. $\mu_{t}^k \in [0, 1]$ is the momentum term for the $k$-th client. When $\mu_{t}^k=0$, the ``HPM" only keeps the current personalized model; when $\mu_{t}^k=1$, the ``HPM" degenerates into an independent private model. Because only a fraction of clients is selected in each round, the update frequency and learning speed of clients are slightly distinct. Hence, the momentum should be client-specific and dynamically adjusted. We assign a counter $z_{t}^k$ as the number of times that the $k$-th client has been selected. If $k \in \calS_t$, then $z_{t}^k = z_{t-1}^k + 1$. Then we linearly set $\mu_{t}^k = \mu * z_{t}^k / (Q * T)$, where $Q$ is the client selection ratio, $T$ is the maximum number of global rounds, and $Q * T$ denotes the expected times of being selected. $\mu$ is the macro momentum that controls the change of $\mu_{t}^k$, and we take $\mu=0.9$ in this paper by default. Finally, we limit $\mu_{t}^k$ in a range of $[0, 1]$.
	
	Then, we exploit the ``HPM" to supervise the personalization process of the newly downloaded model in the next communication round. In the first communication round of FL, we do not apply this process because we have not stored the ``HPM". We explore two specific forms of knowledge transfer. For a specific sample $\bfx_i^k$, we denote the outputs of $\psi_{t}^k$ and $\psi_{\text{p},t}^k$ as $\bfg_{i}$ and $\bfg_{\text{p},i}$, respectively; and their intermediate features as $\bfh_{i}$ and $\bfh_{\text{p},i}$, respectively. The outputs are ``logits" without softmax, while the intermediate features are $d$-dimension vectors extracted by the feature extractor. We omit the index of $k$ and $t$ for simplification.
	
	\begin{figure}[tbp]
		\centering
		\includegraphics[width=\linewidth]{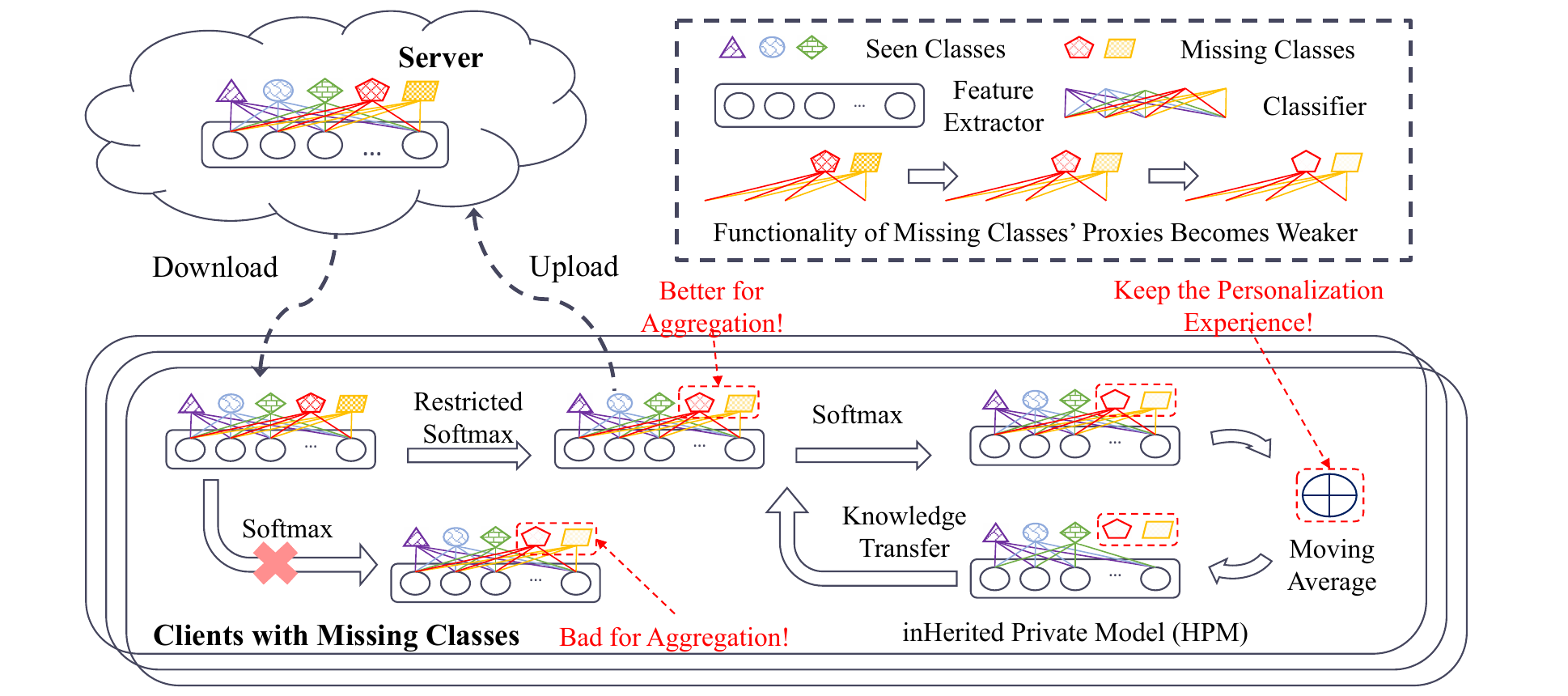}
		\caption{{\small The architecture and procedure of the proposed MAP.}}
		\label{fig:method}
	\end{figure}
	
	First, we could transfer the knowledge contained in the outputs. The knowledge distillation~\cite{KD} could be utilized to enhance the information transfer via:
	\begin{equation}
		\calL_{\text{kd}}\left( \psi_{t}^k \right) = \tau^2 \frac{1}{B} \sum_{i=1}^B D_{KL}\left( \sigma(\bfg_{\text{p},i}/\tau) || \sigma(\bfg_{i}/\tau) \right), \label{eq:kd}
	\end{equation}
	where $D_{KL}$ refers to the KL-divergence, $\sigma(\cdot)$ is the softmax operation, $B$ denotes the batch size, and $\tau$ is the temperature. We fix $\tau=4.0$ which is commonly utilized in knowledge distillation. We can also transfer the knowledge contained in the intermediate features. We can take the maximum mean discrepancy (MMD)~\cite{MMD} to align the feature distributions between the newly-downloaded model and the ``HPM":
	\begin{equation}
		\calL_{\text{mmd}}\left( \psi_{t}^k \right) = \left \lVert \frac{1}{B}\sum_{i=1}^B\Phi(\bfh_{i}) - \frac{1}{B}\sum_{i=1}^B\Phi(\bfh_{\text{p},i}) \right \rVert_{\mathcal{H}}^2, \label{eq:mmd}
	\end{equation}
	where $\Phi(\cdot)$ is a feature map induced by a specific kernel function, i.e., $k(\bfh_i, \bfh_j) = \langle \Phi(\bfh_i), \Phi(\bfh_j) \rangle$. We use multiple Gaussian kernels with different bandwidths as in~\cite{DAN}. With these types of knowledge transfer, and combined with Eq.~\ref{eq:local-target}, the total personalization loss is denoted as:
	\begin{equation}
		\calL_{\text{total}}\left( \psi_{t}^k \right) = (1 - \lambda) \calL\left( \psi_{t}^k \right) + \lambda \calL_{\text{kt}}\left( \psi_{t}^k \right), \label{eq:total-loss}
	\end{equation}
	where $\calL_{\text{kt}}$ could be $\mathcal{L}_{\text{kd}}$ or $\mathcal{L}_{\text{mmd}}$. $\calL_{\text{kt}}$ and $\lambda$ are hyper-parameters. The personalized model $\hat{\psi}_t^k$ could be obtained via several steps of SGD update as $\psi_t^k \leftarrow \psi_t^k - \eta \nabla_{\psi_{t}^k} \calL_{\text{total}}\left( \psi_{t}^k \right)$. $\eta$ is the learning rate.
	
	\subsubsection{Theoretical Analysis}
	We provide a macroscopic analysis of the advantages of ``HPM". Similar to FedBoost~\cite{FedBoost}, we utilize the Bregman Divergence $B_{F}$ as the loss function and assume that $F$ is strictly convex and $B_{F}$ is jointly convex. The loss of the $k$-th client is $ B_F\left(h^k || \mathcal{D}^k \right) = F(h^k) - F(\mathcal{D}^k) - \left\langle \nabla F(\mathcal{D}^k), h^k - \mathcal{D}^k \right\rangle $, where $h^k$ is the learned estimator with a little abuse of notations.
	The updated ``HPM" in the $t$-th round is actually an interpolation as: $h_{\text{p}, t+1}^{k} = (1-\alpha) \hat{h}_{t}^{k} + \alpha h_{\text{p}, t}^{k}$, with $\alpha = \mu_t^k$. According to the property of Bregman Divergence, the local loss function is bounded:
	\begin{equation}
		\nonumber
		B_F\left( h_{\text{p}, t+1}^{k} || \calD^{k} \right) \leq (1-\alpha) B_F\left( \hat{h}_{t}^{k} || \calD^{k} \right) + \alpha B_F\left( h_{\text{p}, t}^{k} || \calD^{k} \right), \label{eq:fedphp-bound}
	\end{equation}
	which bounds the personalization error via two components: $B_F\left( \hat{h}_{t}^k || \calD^k \right)$ denotes the error of after personalizing $h_{t}^k$, which can be small with appropriate fine-tuning; $B_F\left(h_{\text{p}, t}^{k} || \calD^k \right)$ denotes the error of the $t$-th ``HPM", which can be deduced similarly to the $(t-1)$-th round.
	To be brief, the macroscopic theoretical analysis shows that ``HPM" is a special kind of private-shared model that can inherit the historical personalized models' ability, leading to smaller personalization errors.

	\begin{algorithm}[tb]
		\caption{{Model Aggregation and Personalization (MAP)}}
		\label{algo:map}
		\textbf{HyperParameters}: $Q$, $T$, $E$, $B$, $\alpha$ in Eq.~\ref{eq:rs}; $\lambda$ and $\mathcal{L}_{\text{kt}}$ in Eq.~\ref{eq:total-loss} \\
		\textbf{Return}: $\psi$: the final aggregated global model;  $\{\psi^k\}_{k=1}^K$: the personalization model for each local client \\
		\textbf{ServerProcedure}:
		\begin{algorithmic}[1]
			\FOR{global round $t = 1, 2, \ldots, T$}
			\STATE $S_t \leftarrow $ sample $\max(Q \cdot K, 1)$ clients
			\FOR{$k \in S_t$}
			\STATE $\hat{\psi}_{\text{a},t}^k, \psi_{\text{p},t+1}^k \leftarrow $ ClientProcedure($k$, $\psi_{t}$)
			\ENDFOR
			\STATE $\psi_{t+1} \leftarrow \sum_{k \in S_t}\frac{1}{|S_t|} \hat{\psi}_{\text{a},t}^k$
			\ENDFOR
			\STATE \textbf{Return}: $\psi \leftarrow \psi_{T + 1}$, and $\psi^k \leftarrow \psi_{\text{p},T+1}^k, \forall k \in [K]$
		\end{algorithmic}
		\textbf{ClientProcedure}($k$, $\psi_{t}$):
		\begin{algorithmic}[1]
			\STATE $\psi_{t}^k \leftarrow \psi_{t}$
			\FOR{local epoch $e = 1, 2, \ldots, E / 2$}
			\FOR{each batch $\left\{\left(\bfx_{i}^k, y_{i}^k\right)\right\}_{i=1}^B$ sampled from $\mathcal{D}_k$}
			\STATE Update $\psi^k_{t}$ via the ``RS" (Eq.~\ref{eq:rs})
			\ENDFOR
			\ENDFOR			
			\STATE Obtain the model for aggregation: $\hat{\psi}_{\text{a},t}^k \leftarrow \psi_{t}^k$
			\FOR{local epoch $e = E / 2 + 1, E / 2 + 2, \ldots, E$}
			\FOR{each batch $\left\{\left(\bfx_{i}^k, y_{i}^k\right)\right\}_{i=1}^B$ sampled from $\mathcal{D}_k$}
			\STATE Update $\psi^k_{t}$ via the softmax and the KD loss (Eq.~\ref{eq:total-loss})
			\ENDFOR
			\ENDFOR
			\STATE Obtain the personalized model: $\hat{\psi}_{t}^k \leftarrow \psi_{t}^k$
			\STATE Update the ``HPM": $\psi_{\text{p},t+1}^k \leftarrow (1-\mu_{t}^k) \hat{\psi}_{t}^k + \mu_t^k \psi_{\text{p},t}^k $
			\STATE Adjust $\mu_t^k$ as in Sect.~\ref{sect:hpm-per}
			\STATE \textbf{Return}: $\hat{\psi}_{\text{a},t}^k$, $\psi_{\text{p},t+1}^k$
		\end{algorithmic}
	\end{algorithm}

	\subsection{Proposed MAP and Discussions}
	We propose MAP via seamlessly combining the proposed ``RS" and ``HPM" in FL. The illustration is shown in Fig.~\ref{fig:method}. We follow the architecture and training paradigm of FedAvg~\cite{FedAvg}. Different from FedAvg, we take two stages of training in each personalization procedure. In the first personalization stage, we utilize ``RS" (Eq.~\ref{eq:rs}) instead of normal softmax to restrict the update of missing classes' proxies, and the updated model is sent to the server for better aggregation. In the second personalization stage, we switch ``RS" to softmax and continue training on the client's local data. The personalization is supervised via the normal training loss and the knowledge distillation loss from the ``HPM" (Eq.~\ref{eq:total-loss}). The final personalized model is fused into the ``HPM" via moving average (Eq.~\ref{eq:move-average}). Obviously, the first stage benefits the model aggregation, and the second stage benefits the model personalization. That is, the proposed MAP could simultaneously achieve the model aggregation and personalization goals in FL with incomplete classes. The pseudo-code is listed in Algo.~\ref{algo:map}.

	Discussions from several other aspects could also explain the advantages of the proposed method. Adding scaling factors to softmax can be seen as applying an effective learning rate of $\alpha \eta$, $\alpha \in [0, 1]$~\cite{TempCheck}. 
	``RS" is similar to PC-Softmax~\cite{PC-Softmax}, which corrects the problem of imbalanced data from the aspect of ``scores" as $p_{c} \propto \exp ( p(y=c) \bfw_c^T \bfh )$, where we take a smooth prior distribution $p(y=c) \propto \calI\{c \in \calO\} + \alpha \calI\{c \in \calM\}$.
	The moving average in ``HPM" is inherently one type of temporal ensembling~\cite{TemporalEnsemble}, which works similarly as the self-ensembling and mean teacher in~\cite{MeanTeacher}.

	\section{Experiments}
	We investigate our methods on several FL benchmarks constructed from FaMnist~\cite{FaMnist}, CIFAR-10/100~\cite{cifar}, and CINIC-10~\cite{Cinic10}. We also explore the FeMnist benchmark recommended by LEAF~\cite{LEAF}. For visualization and analysis, we also utilize Mnist~\cite{mnist} data.
	
	\subsection{Observations}
	\subsubsection{Softmax with Missing Classes}
\begin{figure}[tbp]
		\centering
		\includegraphics[width=\linewidth]{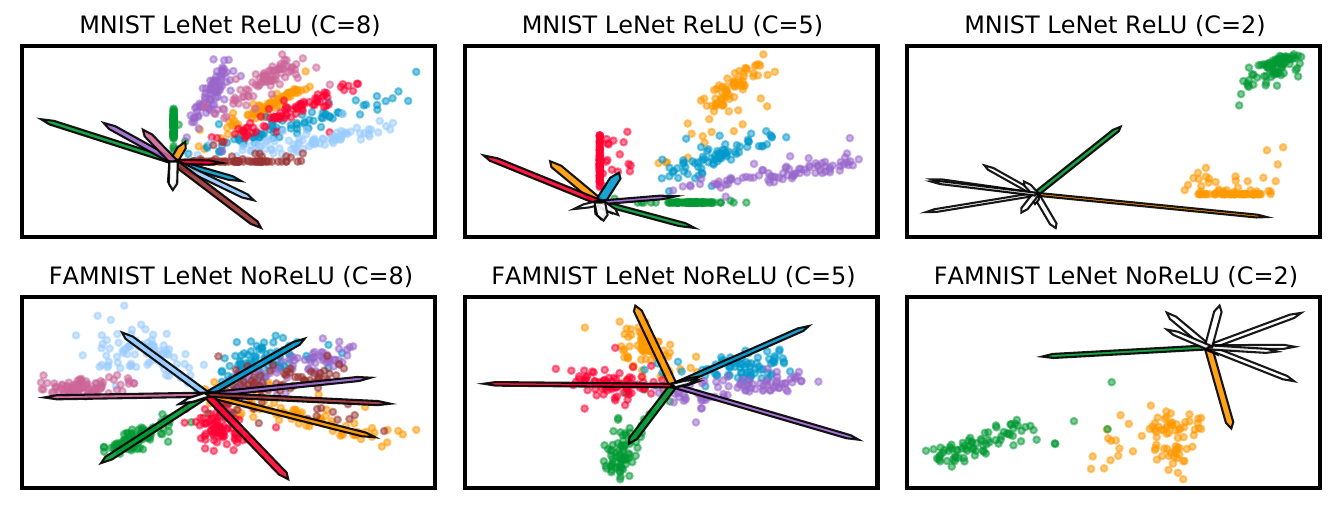}
		\caption{{\small Visualization of learned features and proxies with 8, 5, and 2 observed classes, respectively.}}
		\label{fig:vis-mnist-clf}
	\end{figure}

	We first verify the Property~\ref{prop:softmax-miss-class} via visualization on Mnist. We slightly modify LeNet~\cite{LeNet} by setting the final hidden dimension as 2 for better visualization. We select Mnist data from the first 8, 5, and 2 classes, and then train LeNet on these data correspondingly. We use SGD with a momentum of 0.9 as the optimizer. The learning rate is 0.01. The number of training epochs is 50. The results are shown in Fig.~\ref{fig:vis-mnist-clf}. The top/bottom shows the results of LeNet with/without ReLU activation~\cite{relu} before the final classification layer. Arrows show the learned proxies and the white ones show proxies of missing classes. We scale the norm of proxies by 10x for better visualization. Not so rigorously, the proxies of missing classes are updated towards the negative direction of existing samples' center according to Eq.~\ref{eq:update-miss-w}. First, proxies of missing classes are sometimes squeezed towards the point of origin. Second, as the number of missing classes increases, the features of observed classes become less compact. Regardless of the ReLU activation function, the missing classes' proxies tend to be inaccurate and have a smaller norm.
	
	\begin{figure}[tbp]
		\centering
		\includegraphics[width=\linewidth]{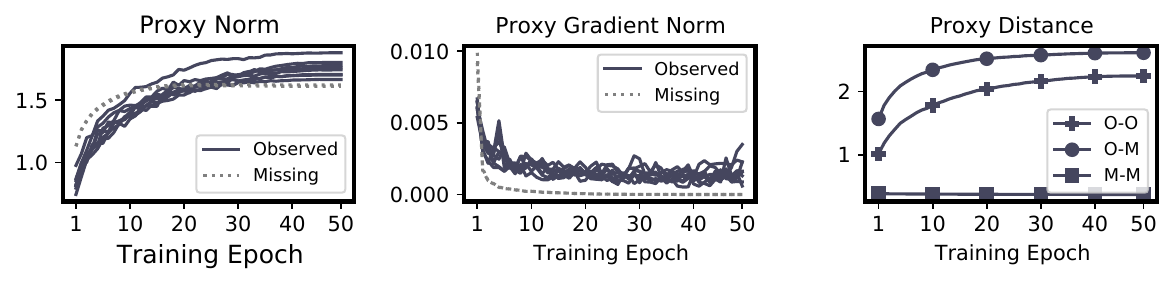}
		\caption{{\small The change of observed and missing classes' proxies during training with incomplete classes.}}
		\label{fig:obser-stats}
	\end{figure}
	
	We also explore the change of proxies during the training process. Specifically, we train VGG11~\cite{VGG} with batch normalization~\cite{bn} on CIFAR-10. We sample the first 8 classes as observed classes and the other 2 classes as missing ones. We still use SGD with a momentum of 0.9 as the optimizer. The learning rate is 0.03. The number of training epochs is 50. After each training epoch, we record some statistics of the proxies including: (1) the norm of these proxies, i.e., $\lVert \mathbf{w}_c \rVert_2$; (2) the norm of these proxies' gradient, i.e., $\lVert \nabla_{\mathbf{w}_c} \calL \rVert_2$, where the gradient is accumulated across this training epoch; (3) the average distance between proxies, i.e., $\sum_{i,j \in \calO} \lVert \mathbf{w}_i - \mathbf{w}_j \rVert_2 / |\calO|^2$,  $\sum_{i \in \calO, j \in \calM} \lVert \mathbf{w}_i - \mathbf{w}_j \rVert_2 / (|\calO|*|\calM|)$, and $\sum_{i,j \in \calM} \lVert \mathbf{w}_i - \mathbf{w}_j \rVert_2 / |\calM|^2$, which are denoted as ``O-O", ``O-M", and ``M-M", respectively. The change of these statistics is shown in Fig.~\ref{fig:obser-stats}. From this figure, we can observe that the norm of missing classes' proxies is slightly smaller than the observed classes, which agrees with the visualization result in Fig.~\ref{fig:vis-mnist-clf}. Then, the gradient norm of missing classes' proxies is large in the beginning and becomes nearly zero after convergence. This implies that the pushing force is stronger at the beginning, which could make the missing classes' proxies lose their effectiveness rapidly. Additionally, as shown in the right of Fig.~\ref{fig:obser-stats}, the distance between missing classes' proxies becomes nearly zero even after only one training epoch. That is, the missing classes' proxies will have no distinctness due to the lack of pulling forces. These experimental observations reveal the disadvantages of model training with incomplete classes.
	
	\begin{figure}[tbp]
		\centering
		\includegraphics[width=\linewidth]{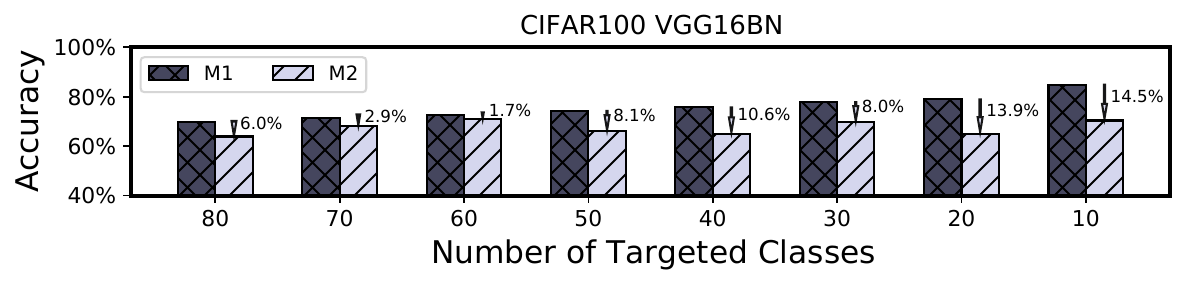}
		\caption{{\small The performance degradation on incomplete classes.}}
		\label{fig:obser-miss-accs}
	\end{figure}
	
	\subsubsection{Performance Degradation on Incomplete Classes}
	The above shows that training with incomplete classes could make the missing classes' proxies inaccurate rapidly, which challenges the goal of building a complete classification model. Then, we want to present that training with all classes' samples may hurt the ability to distinguish a definite set of partial classes.
	
	\begin{table*}[tbp]
		\renewcommand{\arraystretch}{1.1}
		\centering
		\caption{{\small Aggregation performance comparisons with standard FL algorithms ($T=150$, $Q=0.2$).}} \label{tab:compare-agg}
		{
			\begin{tabular}{l|c|c|c|c|c|c|c|c|c}
				\hline
				& \multicolumn{2}{c|}{FaMnist ($E=5$)} & \multicolumn{2}{c|}{CIFAR-10 ($E=10$)} & \multicolumn{2}{c|}{CIFAR-100 ($E=20$)} & \multicolumn{2}{c|}{CINIC-10 ($E=40$)} & \multirow{2}{*}{Avg} \\ \cline{2-9}
				& MLPNet & LeNet & TFCNN & VGG8 & VGG8-BN & VGG11 & ResNet20 & ResNet56 & \\
				\hline
				FedAvg~\cite{FedAvg} & 86.2 {\scriptsize (0.92\%)} & 86.3 {\scriptsize (0.58\%)} & 66.8 {\scriptsize (0.64\%)} & 73.0 {\scriptsize (0.99\%)} & 47.6 {\scriptsize (0.78\%)} & 24.8 {\scriptsize (0.50\%)} & 61.5 {\scriptsize (0.51\%)} & 65.8 {\scriptsize (1.03\%)} & 64.0 \\ \hline
				FedProx~\cite{FedProx} & 86.8 {\scriptsize (0.80\%)} & 87.4 {\scriptsize (0.47\%)} & 70.7 {\scriptsize (0.86\%)} & 74.6 {\scriptsize (0.26\%)} & 47.0 {\scriptsize (0.66\%)} & 24.8 {\scriptsize (0.08\%)} & 64.8 {\scriptsize (1.19\%)} & 66.5 {\scriptsize (0.79\%)} & 65.3 \\ \hline
				FedMMD~\cite{FedMMD} & 87.0 {\scriptsize (4.42\%)} & 88.2 {\scriptsize (0.70\%)} & 69.9 {\scriptsize (1.07\%)} & 74.7 {\scriptsize (0.66\%)} & 46.8 {\scriptsize (0.69\%)} & 26.9 {\scriptsize (0.96\%)} & 64.3 {\scriptsize (1.07\%)} & 66.1 {\scriptsize (0.19\%)} & 65.5 \\ \hline
				Scaffold~\cite{Scaffold} & 88.6 {\scriptsize (0.46\%)} & 89.2 {\scriptsize (0.33\%)} & 73.9 {\scriptsize (0.54\%)} & 76.2 {\scriptsize (0.69\%)} & 52.3 {\scriptsize (0.60\%)} & 21.3 {\scriptsize (0.42\%)} & 65.3 {\scriptsize (0.72\%)} & 66.1 {\scriptsize (0.79\%)} & 66.6 \\ \hline
				FedDF~\cite{FedDF} & 86.6 {\scriptsize (1.94\%)} & 87.5 {\scriptsize (1.05\%)} & 69.1 {\scriptsize (2.06\%)} & 72.6 {\scriptsize (3.56\%)} & 44.0 {\scriptsize (1.87\%)} & 26.9 {\scriptsize (1.41\%)} & 65.1 {\scriptsize (1.42\%)} & 67.2 {\scriptsize (2.48\%)} & 64.9 \\ \hline
				FedOpt~\cite{FedOpt} & 86.3 {\scriptsize (0.12\%)} & 87.6 {\scriptsize (1.36\%)} & 65.7 {\scriptsize (1.46\%)} & 69.1 {\scriptsize (1.26\%)} & 43.0 {\scriptsize (1.40\%)} & 23.6 {\scriptsize (2.18\%)} & 60.4 {\scriptsize (1.33\%)} & 61.7 {\scriptsize (1.89\%)} & 62.2 \\ \hline
				FedNova~\cite{FedNova} & 86.5 {\scriptsize (1.17\%)} & 86.8 {\scriptsize (0.56\%)} & 64.0 {\scriptsize (1.62\%)} & 68.7 {\scriptsize (1.66\%)} & 42.5 {\scriptsize (1.17\%)} & 23.4 {\scriptsize (1.10\%)} & 61.0 {\scriptsize (1.09\%)} & 60.4 {\scriptsize (1.92\%)} & 61.7 \\ \hline
				FedAwS~\cite{FedAwS} & 87.4 {\scriptsize (0.36\%)} & 87.8 {\scriptsize (0.43\%)} & 67.9 {\scriptsize (0.56\%)} & 73.9 {\scriptsize (0.79\%)} & 47.1 {\scriptsize (0.82\%)} & 25.5 {\scriptsize (0.46\%)} & 64.5 {\scriptsize (1.23\%)} & 65.8 {\scriptsize (0.76\%)} & 65.0 \\ \hline
				MOON~\cite{MOON} & 87.4 {\scriptsize (0.14\%)} & 87.8 {\scriptsize (0.41\%)} & 70.6 {\scriptsize (0.73\%)} & 74.6 {\scriptsize (0.83\%)} & 47.1 {\scriptsize (0.53\%)} & 25.1 {\scriptsize (0.15\%)} & 62.1 {\scriptsize (1.13\%)} & 65.9 {\scriptsize (1.19\%)} & 65.1 \\ \hline
				FedDyn~\cite{FedDyn} & 84.9 {\scriptsize (1.63\%)} & 86.8 {\scriptsize (1.08\%)} & 62.0 {\scriptsize (2.18\%)} & 62.4 {\scriptsize (1.02\%)} & 37.4 {\scriptsize (1.58\%)} & 23.0 {\scriptsize (1.62\%)} & 61.1 {\scriptsize (0.76\%)} & 58.4 {\scriptsize (1.52\%)} & 59.5 \\ \hline
				FLDA~\cite{FLDA} & 87.4 {\scriptsize (0.16\%)} & 86.2 {\scriptsize (0.83\%)} & 59.3 {\scriptsize (1.18\%)} & 74.2 {\scriptsize (0.42\%)} & 44.8 {\scriptsize (0.78\%)} & 28.2 {\scriptsize (0.69\%)} & 68.2 {\scriptsize (0.73\%)} & 64.9 {\scriptsize (0.91\%)} & 64.2 \\ \hline
				pFedMe~\cite{pFedMe} & 87.7 {\scriptsize (0.63\%)} & 87.8 {\scriptsize (0.38\%)} & 66.6 {\scriptsize (2.87\%)} & 58.9 {\scriptsize (1.96\%)} & 46.5 {\scriptsize (0.93\%)} & 1.3 {\scriptsize (0.14\%)} & 61.0 {\scriptsize (0.98\%)} & 64.5 {\scriptsize (0.61\%)} & 59.3 \\ \hline
				PerFedAvg~\cite{PerFedAvg} & 87.6 {\scriptsize (4.47\%)} & 88.2 {\scriptsize (4.51\%)} & 70.3 {\scriptsize (3.48\%)} & 67.5 {\scriptsize (3.32\%)} & 45.5 {\scriptsize (1.26\%)} & 1.0 {\scriptsize (0.00\%)} & 64.9 {\scriptsize (0.75\%)} & 63.6 {\scriptsize (0.61\%)} & 61.1 \\ \hline
				Ditto~\cite{Ditto} & 87.5 {\scriptsize (0.60\%)} & 87.5 {\scriptsize (0.89\%)} & 68.4 {\scriptsize (0.52\%)} & 74.2 {\scriptsize (0.95\%)} & 47.3 {\scriptsize (0.29\%)} & 25.3 {\scriptsize (0.40\%)} & 62.8 {\scriptsize (0.49\%)} & 64.7 {\scriptsize (0.86\%)} & 64.7 \\ \hline
				FedROD~\cite{FedROD} & 88.1 {\scriptsize (0.97\%)} & 87.9 {\scriptsize (0.53\%)} & 63.9 {\scriptsize (1.07\%)} & 73.8 {\scriptsize (1.11\%)} & 47.6 {\scriptsize (1.74\%)} & 25.8 {\scriptsize (0.77\%)} & 67.6 {\scriptsize (0.48\%)} & 67.4 {\scriptsize (0.39\%)} & 65.3 \\ \hline
				FedRS~\cite{FedRS} & 87.8 {\scriptsize (0.46\%)} & 88.1 {\scriptsize (0.58\%)} & 70.7 {\scriptsize (1.11\%)} & 75.9 {\scriptsize (1.19\%)} & 48.5 {\scriptsize (0.86\%)} & 25.4 {\scriptsize (0.71\%)} & 67.4 {\scriptsize (1.20\%)} & 69.4 {\scriptsize (1.25\%)} & 66.6 \\ \hline
				FedPHP~\cite{FedPHP} & 87.7 {\scriptsize (0.60\%)} & 87.8 {\scriptsize (0.38\%)} & 70.3 {\scriptsize (0.85\%)} & 74.3 {\scriptsize (0.31\%)} & 47.4 {\scriptsize (0.05\%)} & 26.9 {\scriptsize (0.79\%)} & 63.2 {\scriptsize (1.27\%)} & 66.2 {\scriptsize (0.73\%)} & 65.5 \\
				\hline
				DittoRS & 87.8 {\scriptsize (0.37\%)} & 88.3 {\scriptsize (1.00\%)} & 69.7 {\scriptsize (1.29\%)} & 74.9 {\scriptsize (1.14\%)} & 47.3 {\scriptsize (1.18\%)} & 25.0 {\scriptsize (0.78\%)} & 65.2 {\scriptsize (0.90\%)} & 66.8 {\scriptsize (0.99\%)} & 65.6 \\ \hline
				ScaRS & 88.0 {\scriptsize (1.90\%)} & 89.1 {\scriptsize (1.39\%)} & 74.2 {\scriptsize (2.12\%)} & 75.7 {\scriptsize (1.89\%)} & 54.4 {\scriptsize (1.75\%)} & 21.0 {\scriptsize (1.21\%)} & 62.6 {\scriptsize (0.91\%)} & 66.3 {\scriptsize (0.83\%)} & 66.4 \\
				\hline
				\rowcolor{gray!20} MAP (Ours) & 88.1 {\scriptsize (0.54\%)} & 88.4 {\scriptsize (0.44\%)} & 71.8 {\scriptsize (0.84\%)} & 76.4 {\scriptsize (0.72\%)} & 48.7 {\scriptsize (0.70\%)} & 25.5 {\scriptsize (0.77\%)} & 67.9 {\scriptsize (1.03\%)} & 67.9 {\scriptsize (0.80\%)} & 66.8 \\ \hline
				\rowcolor{gray!20} ScaMAP (Ours) & 88.3 {\scriptsize (1.25\%)} & 89.6 {\scriptsize (0.91\%)} & 75.6 {\scriptsize (1.39\%)} & 76.5 {\scriptsize (1.79\%)} & 52.6 {\scriptsize (1.70\%)} & 21.9 {\scriptsize (0.48\%)} & 64.9 {\scriptsize (1.36\%)} & 66.1 {\scriptsize (0.93\%)} & 66.9 \\ \hline
			\end{tabular}
		}
	\end{table*}
	
	We train VGG16~\cite{VGG} with batch normalization~\cite{bn} on CIFAR-100. CIFAR-100 contains 100 classes and each class has 500 training samples. We aim to compare the models trained on the whole 100 classes or a partial set of targeted classes. The number of targeted classes ranges from 10 to 80. Take 80 targeted classes as an example, we first train a model on the 80 classes' data, i.e., $80\times 500$ samples. Then, we train a model on all classes with the same number of targeted classes, i.e., using all $100 \times 500$ samples. For ease of description, we denote these two models M1, M2, respectively. For performance comparison, we only calculate the model's test accuracy on targeted classes. We still train models up to 50 epochs. The optimizer is SGD with a momentum of 0.9 and the learning rate is 0.03. Fig.~\ref{fig:obser-miss-accs} shows the results, where the dark bars show the accuracy of M1, and the light bars show the accuracies of M2. M2 sees the same number of targeted samples, while it simultaneously has to distinguish other classes well, which still leads to the degradation of performance on targeted classes. As the number of targeted classes decreases, the performance degradation becomes more serious. That is, if one only aims to distinguish a partial set of classes, introducing other classes' samples may degrade the performance. Correspondingly, in FL with incomplete classes, the aggregated global model on the server does not necessarily satisfy the client's personalization goals. We will detail this more in the following sections.
	
	\begin{figure}[!t]
		\centering
		\includegraphics[width=\linewidth]{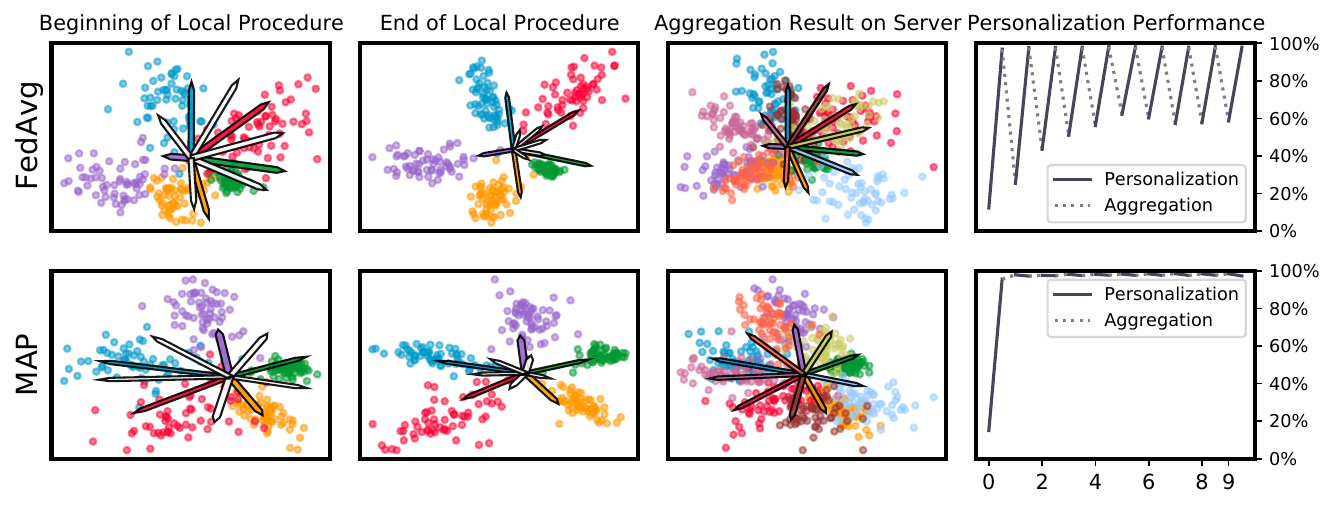}
		\caption{{\small An FL demo that shows the drawbacks of FedAvg (the top) and the advantages of MAP (the bottom).}}
		\label{fig:vis-mnist-fed}
	\end{figure}

	\subsubsection{Visualization in FL with Incomplete Classes}
	Then, we show the drawbacks of softmax and the phenomena of local performance degradation caused by incomplete classes in an FL demo. The simple demo only contains 2 clients, where the first client owns samples from the class set $\{0, 1, 2, 3, 4\}$ while the second one has samples from $\{5, 6, 7, 8, 9\}$. We use LeNet without ReLU activation as the network and set the final dimension as 2 for better visualization. We train models following the procedure of FL. Specifically, we take 10 global communication rounds and train local models up to 5 epochs in each local personalization procedure. We plot the learned features and proxies at the 5th round in Fig.~\ref{fig:vis-mnist-fed}. The two rows compare FedAvg and MAP on the first client. The first column shows the beginning of the local procedure, i.e., the extracted features via the newly-downloaded global model and its proxies. The second column shows the local personalized results with incomplete classes, i.e., only 5 classes. We can obviously observe that proxies of missing classes are suppressed towards zero in FedAvg, while MAP can alleviate this phenomenon owing to the effects of ``RS". Correspondingly, the aggregated proxies become inaccurate in FedAvg, while MAP can get more discriminative proxies in the third column. The last column shows the performance degradation in this simple FL demo. Specifically, we record the local test accuracies of the newly downloaded model and the personalized model in each personalization stage. When using FedAvg, we can observe that the local performances will improve during the personalization procedure (the solid segments), while the newly downloaded models’ performance could be much worse than the last personalized model (the dotted segments). This observation conforms to the motivation in Sect.~\ref{sect:per-obser}. However, MAP could mitigate the performance degradation phenomena with the help of the ``HPM". The above results directly show the problems of traditional FL algorithms faced with incomplete classes and vividly verify the superiorities of MAP.
	
	\subsection{Performance Comparisons}
	\subsubsection{Experimental Settings}
	We then evaluate the aggregation and personalization performances of MAP on several FL benchmarks including FaMnist~\cite{FaMnist}, CIFAR-10/100~\cite{cifar}, and CINIC-10~\cite{Cinic10}. The full name of FaMnist is Fashion-MNIST, which comprises 70,000 images of fashion products from 10 categories. The training set has 60,000 images and the test set has 10,000 images. CIFAR-10 and CIFAR-100 have 10/100 classes to classify. They consist of 50,000 training images and 10,000 test images. CINIC-10 is a combination of CIFAR-10 and ImageNet~\cite{imagenet}, which contains 10 classes. It contains 90,000 samples for training, validation, and testing, respectively. We do not use the validation set. We scale all images to the size of $32\times 32$.
	
	For each benchmark, we vary the utilized neural networks to fully verify the effectiveness of MAP. Specifically, for FaMnist, we utilize MLPNet and LeNet~\cite{LeNet} as the models, respectively. MLPNet contains two hidden layers with each hidden layer containing 512 neurons. Different from the goal of visualization, we utilize the original LeNet without any modifications. For CIFAR-10, we utilize the network used in FedAvg~\cite{FedAvg} (denoted as TFCNN) and VGG8~\cite{VGG}, respectively. We utilize VGG8-BN and VGG11 for CIFAR-100. ResNet20 and ResNet56~\cite{ResNet} are for CINIC-10. The networks are implemented via PyTorch\footnote{\url{https://pytorch.org}}.
	
	Then we introduce the settings of FL experiments. By default, we assume there are 100 clients that are coordinated by one single central server. To simulate the FL scenes with incomplete classes, we randomly assign $c\in\{2,3,\ldots,C\}$ classes to each client and decentralize the training samples of corresponding benchmarks onto these clients. This strategy is similar to several previous works that utilize label partitions to split data~\cite{FedAvg,Fed-NonIID-Data,FedMD}. Different from them, we assume that the number of classes differs among clients. The test samples of corresponding benchmarks are stored on the server to evaluate the aggregation performance. To evaluate the local personalization performance, we randomly leave aside 20\% samples on each client as local test samples. Notably, although the client encounters missing classes, the samples among observed classes are nearly balanced. Hence, we utilize accuracy as the metric for both aggregation and personalization performance evaluation. 
	
	We then report the utilized FL hyper-parameters. We set the number of communication rounds ($T$) as 150 and the client selection ratio ($Q$) as 0.2. That is, we select 20\% clients to participate in the FL procedure in each round. The number of local personalization epochs ($E$) is varied among FaMnist, CIFAR-10, CIFAR-100, and CINIC-10, i.e., 5, 10, 20, and 40 epochs, respectively. Because ResNet could be stably updated with a larger learning rate, we utilize a constant learning rate of 0.03 for the first three benchmarks, and 0.05 for CINIC-10. The optimizer is SGD with a momentum of 0.9. The batch size is 64 and the weight decay is $\text{1e-5}$.
	
	\begin{table*}
		\renewcommand{\arraystretch}{1.1}
		\centering
		\caption{{\small Personalization performance comparisons with standard FL algorithms ($T=150$, $Q=0.2$).}} \label{tab:compare-per}
		{
			\begin{tabular}{l|c|c|c|c|c|c|c|c|c}
				\hline
				& \multicolumn{2}{c|}{FaMnist ($E=5$)} & \multicolumn{2}{c|}{CIFAR-10 ($E=10$)} & \multicolumn{2}{c|}{CIFAR-100 ($E=20$)} & \multicolumn{2}{c|}{CINIC-10 ($E=40$)} & \multirow{2}{*}{Avg} \\ \cline{2-9}
				& MLPNet & LeNet & TFCNN & VGG8 & VGG8-BN & VGG11 & ResNet20 & ResNet56 & \\
				\hline
				FedAvg~\cite{FedAvg} & 90.5 {\scriptsize (0.58\%)} & 91.3 {\scriptsize (0.68\%)} & 73.7 {\scriptsize (0.99\%)} & 77.5 {\scriptsize (0.82\%)} & 47.0 {\scriptsize (0.56\%)} & 25.4 {\scriptsize (0.90\%)} & 82.0 {\scriptsize (0.84\%)} & 82.9 {\scriptsize (0.47\%)} & 71.3 \\ \hline
				FedProx~\cite{FedProx} & 90.7 {\scriptsize (0.51\%)} & 92.0 {\scriptsize (0.79\%)} & 72.0 {\scriptsize (0.76\%)} & 77.2 {\scriptsize (1.18\%)} & 48.4 {\scriptsize (0.80\%)} & 26.4 {\scriptsize (0.76\%)} & 81.2 {\scriptsize (0.40\%)} & 82.1 {\scriptsize (0.52\%)} & 71.2 \\ \hline
				FedMMD~\cite{FedMMD} & 89.8 {\scriptsize (4.53\%)} & 91.2 {\scriptsize (0.39\%)} & 74.0 {\scriptsize (0.99\%)} & 76.5 {\scriptsize (0.55\%)} & 49.4 {\scriptsize (1.01\%)} & 32.6 {\scriptsize (1.25\%)} & 80.7 {\scriptsize (0.27\%)} & 82.9 {\scriptsize (0.80\%)} & 72.1 \\ \hline
				Scaffold~\cite{Scaffold} & 88.7 {\scriptsize (0.42\%)} & 90.3 {\scriptsize (0.13\%)} & 76.1 {\scriptsize (0.71\%)} & 77.9 {\scriptsize (1.04\%)} & 51.2 {\scriptsize (1.16\%)} & 25.7 {\scriptsize (1.11\%)} & 75.5 {\scriptsize (0.74\%)} & 77.6 {\scriptsize (1.24\%)} & 70.4 \\ \hline
				FedDF~\cite{FedDF} & 90.2 {\scriptsize (1.64\%)} & 90.9 {\scriptsize (0.60\%)} & 72.8 {\scriptsize (1.96\%)} & 76.6 {\scriptsize (3.01\%)} & 47.2 {\scriptsize (1.55\%)} & 26.3 {\scriptsize (1.20\%)} & 83.0 {\scriptsize (1.63\%)} & 82.6 {\scriptsize (1.89\%)} & 71.2 \\ \hline
				FedOpt~\cite{FedOpt} & 92.2 {\scriptsize (1.23\%)} & 91.0 {\scriptsize (0.68\%)} & 70.7 {\scriptsize (1.72\%)} & 72.7 {\scriptsize (1.14\%)} & 51.6 {\scriptsize (1.60\%)} & 25.8 {\scriptsize (1.33\%)} & 81.6 {\scriptsize (1.42\%)} & 84.9 {\scriptsize (1.92\%)} & 71.3 \\ \hline
				FedNova~\cite{FedNova} & 89.6 {\scriptsize (1.20\%)} & 90.5 {\scriptsize (0.79\%)} & 68.7 {\scriptsize (1.91\%)} & 71.8 {\scriptsize (1.34\%)} & 43.9 {\scriptsize (1.54\%)} & 25.6 {\scriptsize (1.24\%)} & 79.3 {\scriptsize (1.31\%)} & 79.5 {\scriptsize (1.53\%)} & 68.6 \\ \hline
				FedAwS~\cite{FedAwS} & 89.9 {\scriptsize (0.60\%)} & 90.8 {\scriptsize (0.46\%)} & 72.0 {\scriptsize (0.89\%)} & 76.2 {\scriptsize (0.62\%)} & 51.1 {\scriptsize (1.55\%)} & 29.7 {\scriptsize (1.26\%)} & 82.5 {\scriptsize (0.79\%)} & 81.5 {\scriptsize (0.30\%)} & 71.7 \\ \hline
				MOON~\cite{MOON} & 89.6 {\scriptsize (0.36\%)} & 90.7 {\scriptsize (0.28\%)} & 72.6 {\scriptsize (0.85\%)} & 75.5 {\scriptsize (0.92\%)} & 49.7 {\scriptsize (0.86\%)} & 28.1 {\scriptsize (0.28\%)} & 82.2 {\scriptsize (0.66\%)} & 82.2 {\scriptsize (0.93\%)} & 71.3 \\ \hline
				FedDyn~\cite{FedDyn} & 87.2 {\scriptsize (1.09\%)} & 89.7 {\scriptsize (1.88\%)} & 63.6 {\scriptsize (1.98\%)} & 65.5 {\scriptsize (2.07\%)} & 39.7 {\scriptsize (1.08\%)} & 23.3 {\scriptsize (1.17\%)} & 76.7 {\scriptsize (0.52\%)} & 75.9 {\scriptsize (1.10\%)} & 65.2 \\ \hline
				FedRep~\cite{FedRep} & 90.2 {\scriptsize (0.78\%)} & 87.5 {\scriptsize (0.86\%)} & 61.7 {\scriptsize (1.23\%)} & 67.5 {\scriptsize (0.75\%)} & 32.0 {\scriptsize (0.50\%)} & 19.2 {\scriptsize (0.25\%)} & 81.6 {\scriptsize (0.94\%)} & 81.0 {\scriptsize (0.07\%)} & 65.1 \\ \hline
				FLDA~\cite{FLDA} & 89.9 {\scriptsize (0.44\%)} & 88.3 {\scriptsize (0.15\%)} & 62.2 {\scriptsize (0.70\%)} & 73.1 {\scriptsize (0.95\%)} & 44.2 {\scriptsize (0.92\%)} & 24.6 {\scriptsize (0.67\%)} & 80.4 {\scriptsize (0.88\%)} & 78.1 {\scriptsize (0.86\%)} & 67.6 \\ \hline
				pFedMe~\cite{pFedMe} & 91.2 {\scriptsize (0.55\%)} & 91.5 {\scriptsize (0.85\%)} & 72.5 {\scriptsize (3.10\%)} & 64.0 {\scriptsize (2.16\%)} & 48.8 {\scriptsize (1.15\%)} & 1.7 {\scriptsize (0.35\%)} & 76.5 {\scriptsize (0.87\%)} & 76.3 {\scriptsize (0.56\%)} & 65.3 \\ \hline
				PerFedAvg~\cite{PerFedAvg} & 91.2 {\scriptsize (4.64\%)} & 92.1 {\scriptsize (4.68\%)} & 75.0 {\scriptsize (3.76\%)} & 72.6 {\scriptsize (3.62\%)} & 46.1 {\scriptsize (0.73\%)} & 1.2 {\scriptsize (0.17\%)} & 80.8 {\scriptsize (0.80\%)} & 82.6 {\scriptsize (1.29\%)} & 67.7 \\ \hline
				FedProto~\cite{FedProto} & 88.0 {\scriptsize (0.92\%)} & 88.4 {\scriptsize (1.23\%)} & 59.1 {\scriptsize (2.22\%)} & 59.9 {\scriptsize (2.17\%)} & 26.7 {\scriptsize (0.12\%)} & 21.2 {\scriptsize (1.04\%)} & 67.7 {\scriptsize (0.77\%)} & 63.4 {\scriptsize (1.12\%)} & 59.3 \\ \hline
				Ditto~\cite{Ditto} & 87.0 {\scriptsize (0.71\%)} & 88.2 {\scriptsize (1.11\%)} & 55.3 {\scriptsize (0.21\%)} & 57.9 {\scriptsize (1.64\%)} & 26.0 {\scriptsize (0.72\%)} & 19.0 {\scriptsize (1.10\%)} & 66.0 {\scriptsize (0.18\%)} & 57.4 {\scriptsize (0.59\%)} & 57.1 \\ \hline
				FedROD~\cite{FedROD} & 89.1 {\scriptsize (0.21\%)} & 90.0 {\scriptsize (0.86\%)} & 63.2 {\scriptsize (0.35\%)} & 73.3 {\scriptsize (1.13\%)} & 49.4 {\scriptsize (1.24\%)} & 22.8 {\scriptsize (1.35\%)} & 81.6 {\scriptsize (0.33\%)} & 83.5 {\scriptsize (1.21\%)} & 69.1 \\ \hline
				FedRS~\cite{FedRS} & 90.4 {\scriptsize (1.09\%)} & 91.7 {\scriptsize (1.21\%)} & 72.9 {\scriptsize (1.98\%)} & 77.1 {\scriptsize (1.97\%)} & 50.5 {\scriptsize (1.43\%)} & 30.1 {\scriptsize (1.32\%)} & 81.8 {\scriptsize (1.41\%)} & 82.9 {\scriptsize (1.06\%)} & 72.2 \\ \hline
				FedPHP~\cite{FedPHP} & 90.9 {\scriptsize (0.63\%)} & 92.1 {\scriptsize (0.78\%)} & 77.3 {\scriptsize (0.91\%)} & 80.7 {\scriptsize (0.84\%)} & 50.5 {\scriptsize (1.31\%)} & 33.2 {\scriptsize (1.69\%)} & 83.2 {\scriptsize (0.81\%)} & 83.0 {\scriptsize (0.97\%)} & 73.9 \\
				\hline
				DittoRS & 87.2 {\scriptsize (0.79\%)} & 88.8 {\scriptsize (0.86\%)} & 55.5 {\scriptsize (0.55\%)} & 57.3 {\scriptsize (1.13\%)} & 25.8 {\scriptsize (0.55\%)} & 19.6 {\scriptsize (1.26\%)} & 67.7 {\scriptsize (0.67\%)} & 63.9 {\scriptsize (0.71\%)} & 58.2 \\ \hline
				ScaRS & 88.6 {\scriptsize (1.03\%)} & 90.5 {\scriptsize (1.11\%)} & 77.0 {\scriptsize (1.48\%)} & 78.4 {\scriptsize (1.85\%)} & 53.2 {\scriptsize (2.11\%)} & 21.8 {\scriptsize (0.83\%)} & 75.1 {\scriptsize (1.00\%)} & 77.7 {\scriptsize (0.88\%)} & 70.3 \\
				\hline
				\rowcolor{gray!20} MAP (Ours) & 92.1 {\scriptsize (0.90\%)} & 91.8 {\scriptsize (1.18\%)} & 76.9 {\scriptsize (0.97\%)} & 80.4 {\scriptsize (0.93\%)} & 52.2 {\scriptsize (1.08\%)} & 31.7 {\scriptsize (1.00\%)} & 84.7 {\scriptsize (1.15\%)} & 81.0 {\scriptsize (0.64\%)} & 73.9 \\ \hline
				\rowcolor{gray!20} ScaMAP (Ours) & 90.5 {\scriptsize (0.90\%)} & 91.9 {\scriptsize (0.75\%)} & 81.7 {\scriptsize (1.35\%)} & 81.2 {\scriptsize (1.48\%)} & 53.3 {\scriptsize (1.78\%)} & 25.4 {\scriptsize (1.45\%)} & 82.5 {\scriptsize (1.75\%)} & 82.5 {\scriptsize (0.25\%)} & 73.6 \\ \hline
			\end{tabular}
		}
	\end{table*}
	
	\subsubsection{Compared Methods}
	We compare MAP with popular FL algorithms as follows.
	FedAvg~\cite{FedAvg} is the most standard FL algorithm.
	FedProx~\cite{FedProx}, FedMMD~\cite{FedMMD} and FedDyn~\cite{FedDyn} introduce various regularization techniques during local personalization procedures. We use the public code for FedDyn~\footnote{\url{https://github.com/alpemreacar/FedDyn}}. Scaffold~\cite{Scaffold} introduces control variates to correct the client-drift problems.
	FedDF~\cite{FedDF} uses a public data set for robust model fusion via ensemble distillation. We randomly sample 500 samples from the global test set as the public data set.
	FedOpt~\cite{FedOpt} and FedNova~\cite{FedNova} takes better optimization strategies in FL.
	FedAwS~\cite{FedAwS} originally solves the extreme case where each client only observes one class. The spreadout term could be also applied to the considered FL scene in this paper.
	MOON~\cite{MOON} takes advantage of the contrastive learning method at the model-level.
	FedRep~\cite{FedRep} and FLDA~\cite{FLDA} utilize private-shared models for better personalization in FL.
	pFedMe~\cite{pFedMe} uses Moreau envelopes as clients’ regularized loss for personalization.
	PerFedAvg~\cite{PerFedAvg} applies meta learning to FL for fast model adaptation. We use the public code for pFedMe and PerFedAvg~\footnote{\url{https://github.com/CharlieDinh/pFedMe}}.
	FedProto~\cite{FedProto} shares class prototypes instead of gradients for federated personalization.
	Ditto~\cite{Ditto} is the variant of FedProx~\cite{FedProx} for federated personalization.
	FedROD~\cite{FedROD} combines class imbalanced loss and private-shared model for both aggregation and personalization. We use the linear version of FedROD and set the final classification layer as the G-Head and P-Head.
	FedRS~\cite{FedRS} is one of the precursors to MAP, which introduces ``RS" for better model aggregation.
	FedPHP~\cite{FedPHP} is another precursor to MAP, which introduces ``HPM" for better model personalization.
	
	These methods almost cover all of the mainstream approaches that are designed to tackle Non-I.I.D. problems in FL. Our method aims to simultaneously achieve model aggregation and personalization goals in FL, and hence, we name our method MAP. The ``RS" and ``HPM" in MAP could also be combined with other FL algorithms. For example, we combine MAP with Scaffold~\cite{Scaffold} and obtain ScaMAP. We run each compared FL method 3 times on average with varied hyper-parameters, and the best performances are reported. The hyper-parameters of MAP include the $\alpha$ in ``RS" (Eq.~\ref{eq:rs}), and $\calL_{\text{kt}},\lambda$ in ``HPM" (Eq.~\ref{eq:total-loss}). We will study the hyper-parameters in the following experimental studies. The default setting is $\alpha=0.9$, $\calL_{\text{kt}}=\calL_{\text{kd}}$, $\lambda=0.01$.
	
	\subsubsection{Results and Analysis}
	We show the aggregation and personalization performances in Tab.~\ref{tab:compare-agg} and Tab.~\ref{tab:compare-per}, respectively. The last column shows the average of the previous eight columns. For aggregation performances, we record the test accuracy of the finally aggregated model on the global test set after $T=150$ communication rounds. For personalization performances, we first calculate the test accuracy on the local test set after personalization for each participated client individually and then report their average accuracy. Notably, we use the personalized model instead of the newly-downloaded model to evaluate the local personalization performance, which is similar to the ``Finetune'' method in previous works~\cite{FedPM}. Commonly, the local test accuracy could improve a lot after personalization, e.g., the demo shown in the last column of Fig.~\ref{fig:vis-mnist-fed}. That is, the personalization performances of compared algorithms are stronger baselines. The metrics are calculated based on Definition~\ref{def-aggregation} and Definition~\ref{def-personalization}, respectively, where we only report the results in the last communication round. We vary several groups of hyper-parameters for investigated FL algorithms and report the best performance among these settings. Additionally, the standard deviations under these hyper-parameter settings are also reported. To compare the FL algorithms more intuitively, the average results of the $4 \times 2 = 8$ experimental groups (i.e., 4 benchmarks and 2 network architectures for each benchmark) are listed in the last column. Most FL algorithms perform well on FaMnist. CIFAR-100 is a harder benchmark that needs to identify 100 classes. Especially, utilizing VGG11 without batch normalization makes the scene harder. Because the clients only need to identify a partial set of classes, the personalization performances are commonly higher than the aggregation performances.

	As shown in Tab.~\ref{tab:compare-agg}, the aggregation performance of FedAvg is about $64.0\%$ on average. Some advanced FL algorithms could indeed improve the aggregation performances, e.g., FedProx obtains $65.3\%$, MOON obtains $65.1\%$. Among compared FL algorithms, Scaffold and FedRS obtain the best aggregation result, i.e., $66.6\%$. FedRep and FedProto are only designed for personalization and do not keep a complete aggregated model on the server, and we set their aggregation performance as $0.0\%$. Our proposed MAP and ScaMAP could surpass FedRS and Scaffold slightly, which obtain $66.8\%$ and $66.9\%$, respectively. 	
	Tab.~\ref{tab:compare-per} shows the personalization performances. FedAvg achieves a personalization performance of $71.3\%$ on average. First, the algorithms that are customized for aggregation may perform worse, e.g., Scaffold only obtains $70.4\%$. One exception is FedRS, which obtains $72.2\%$ on average. Some FL personalization methods could perform much worse than FedAvg, e.g., pFedMe obtains $65.3\%$ and PerFedAvg obtains $67.7\%$. The reasons are that these algorithms may have more than three hyper-parameters to tune. Under the consideration of limited computation resources and fair comparison, we only vary several groups of hyper-parameters. If we enlarge the search space of their hyper-parameters, these FL personalization algorithms may show better performances. Among of the compared methods, FedPHP achieves the best personalization result, i.e., $73.9\%$. Our proposed MAP and ScaMAP could also obtain comparable results, i.e., $73.9\%$ and $73.6\%$, respectively.
	
	\begin{figure}[!t]
		\centering
		\includegraphics[width=\linewidth]{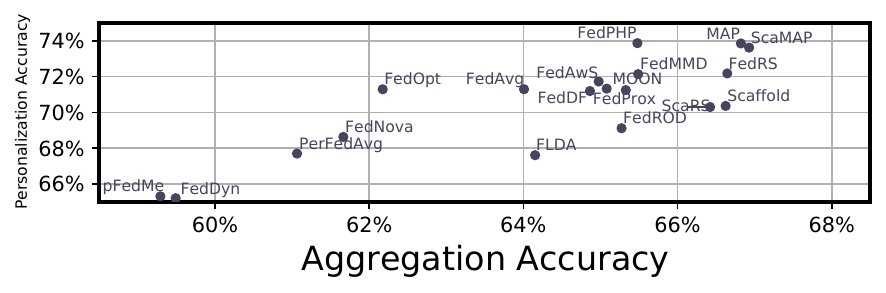}
		\caption{{\small The joint look of aggregation and personalization results.}}
		\label{fig:joint-compare}
	\end{figure}
	
	For fair comparison under incomplete classes, we also apply the proposed ``RS" to Ditto~\cite{Ditto} and Scaffold~\cite{Scaffold}, which are denoted as DittoRS and ScaRS, respectively. From Tab.~\ref{tab:compare-agg} and Tab.~\ref{tab:compare-per}, DittoRS could improve the aggregation and personalization performances of Ditto, i.e., $64.7\%$ to $65.6\%$, and $57.1\%$ to $58.2\%$, respectively. This implies the effectiveness of the proposed ``RS". However, the personalization accuracy is still weaker than FedPHP and MAP. Scaffold could already obtain good aggregation performances and ScaRS does not improve its aggregation ability. Compared with ScaRS, ScaMAP additionally utilizes the proposed ``HPM" for personalization. Hence, the personalization accuracy of ScaMAP is significantly higher than Scaffold and ScaRS, which verifies the effectiveness of the proposed ``HPM".
	
	Considering the aggregation performance, the proposed MAP and ScaMAP only show slightly better results than Scaffold and FedRS. Considering the personalization performance, the proposed MAP and ScaMAP achieve comparable results as the FedPHP. It seems that our proposed methods do not show obvious advantages. We then combine the aggregation and personalization accuracies of each FL algorithm (i.e., the last columns of Tab.~\ref{tab:compare-agg} and Tab.~\ref{tab:compare-per}), and the scatter plot is shown in Fig.~\ref{fig:joint-compare}. Obviously, the FL algorithms in the upper right could simultaneously obtain better aggregation and personalization results. Although Scaffold and FedRS basically stay on the rightmost of the figure, they are not very near the top edge (i.e., a lower personalization performance). Similarly, although FedPHP basically stays on the topmost of the figure, it is not very near the right edge (i.e., a lower aggregation performance). Our methods could also perform better than the recently proposed FedROD which considers both aggregation and personalization in FL. From the above analysis, we conclude that MAP and ScaMAP could simultaneously obtain better aggregation and personalization accuracies, which is the core superiority towards other FL algorithms.
	
	\begin{figure}[!tbp]
		\centering
		\includegraphics[width=\linewidth]{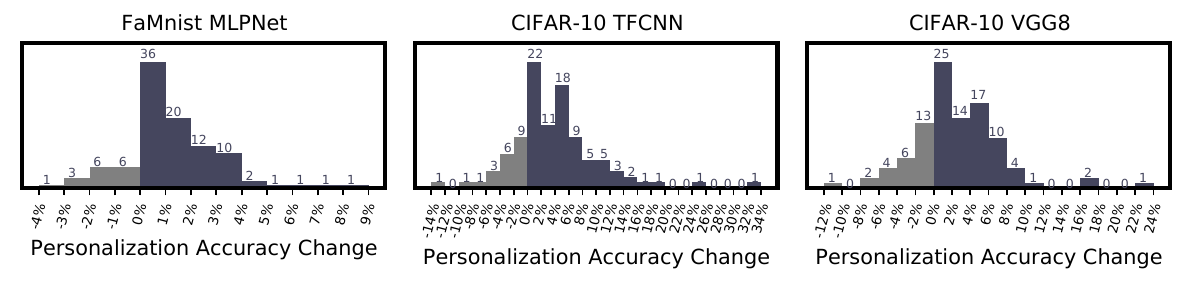}
		\caption{{\small The personalization accuracy change of MAP compared with FedAvg.}}
		\label{fig:fair}
	\end{figure}
	
	\begin{figure}[!tbp]
		\centering
		\includegraphics[width=\linewidth]{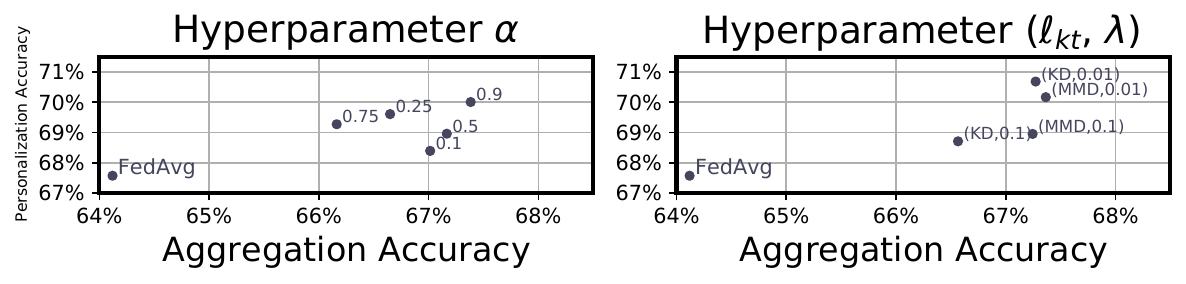}
		\caption{{\small The hyper-parameter studies of our proposed methods.}}
		\label{fig:hypers}
	\end{figure}
	
	\begin{figure}[!t]
		\centering
		\includegraphics[width=\linewidth]{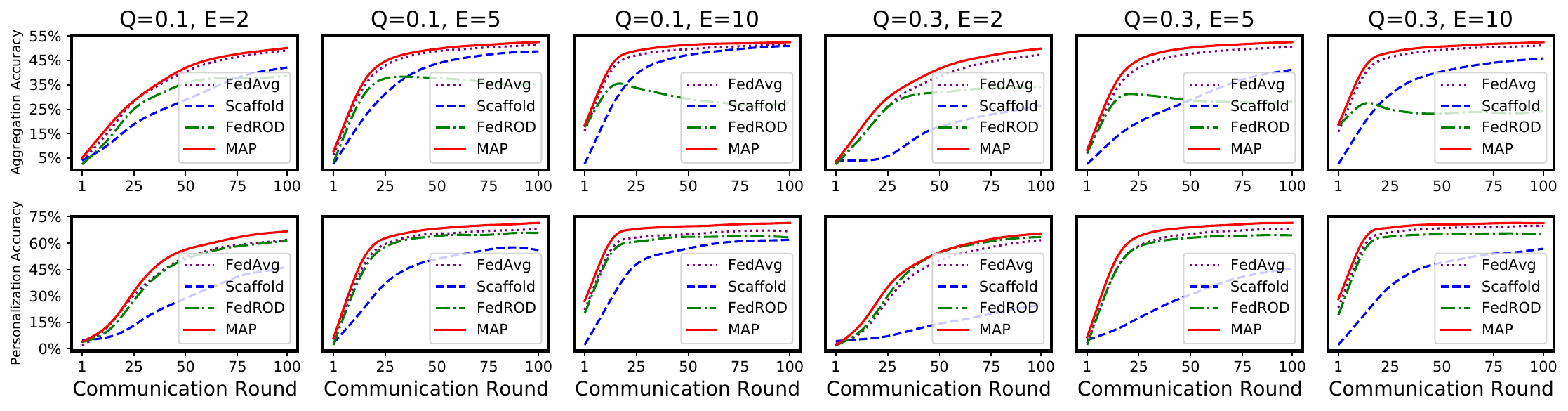}
		\caption{{\small Comparisons of convergence curves on FEMNIST.}}
		\label{fig:performance-more}
	\end{figure}
	
	\subsubsection{Personalization Fairness Comparisons}
	Although using the average accuracy to measure personalization performance is common, the fairness among clients is also significant. Fig.~\ref{fig:fair} shows the change in client personalization accuracy from FedAvg to MAP, where the histogram of 100 clients is calculated and the number of clients is displayed on the bars. Clearly, most of the clients could obtain improvement from MAP, e.g., only 16 clients meet performance degradation on FaMnist. How to further ensure that these clients can also obtain performance improvement is future research work.
	
	\subsubsection{Hyper-Parameter Studies}
	We then study the hyper-parameters of MAP, i.e., the $\alpha$ in ``RS" (Eq.~\ref{eq:rs}), and $\calL_{\text{kt}},\lambda$ in ``HPM" (Eq.~\ref{eq:total-loss}). Specifically, we vary the hyper-parameters of MAP and record the performances like the Tab.~\ref{tab:compare-agg} and Tab.~\ref{tab:compare-per}. We do not record the results on CINIC-10 because running ResNet on it costs too much. That is, we only record the performances of $3 \times 2 = 6$ experimental groups (i.e., 3 benchmarks and 2 network architectures for each benchmark) and plot the scatter of the average results. The space of hyper-parameter is $\alpha \in \{0.1, 0.25, 0.5, 0.75, 0.9\}$, and $(\calL_{\text{kt}},\lambda) \in \{(\calL_{\text{kd}}, 0.01), (\calL_{\text{kd}}, 0.1), (\calL_{\text{mmd}}, 0.01), (\calL_{\text{mmd}}, 0.1)\}$.  Fig.~\ref{fig:hypers} shows the comparison among different hyper-parameters. Setting $\alpha=0.9$, and $\calL_{\text{kt}} = \calL_{\text{kd}}, \lambda=0.01$ is preferred. Although the performance under different hyper-parameters varies, the results are all superior to the most standard FL algorithm, i.e., FedAvg. That is, our methods could uniformly surpass FedAvg under various hyper-parameters.
	
	\subsection{Large-Scale Datasets}
	We also investigate the performances on large-scale FL scenes, i.e., the FEMNIST recommended by LEAF\footnote{\url{https://leaf.cmu.edu/}}~\cite{LEAF}. There are 3550 different users in FEMNIST and each user owns 229 samples on average. We split local data into $80\%$ as the local training set on this client and the other $20\%$ as the local test set. The local test set across all clients are combined as a global test set. Although the scene is not a typical FL scene with incomplete classes, there exists an obvious class-missing phenomenon in each client's local data. Specifically, FEMNIST contains 62 classes in total while each client could only observe 25 classes on average whose sample size is greater than 2. Hence, for each local client, we view classes with samples less than 2 in FEMNIST as missing classes. We train FL algorithms with different $Q$ and $E$ on FEMNIST. The number of communication rounds is 100. The client selection ratio ($Q$) is varied in $\{0.1, 0.3\}$, while the number of local personalization epochs ($E$) is varied in $\{2, 5, 10\}$. We use a network with two convolution layers and one fully-connected layer. We use SGD with a momentum of 0.9, and the learning rate is $0.004$. For better visualization, we only compare MAP with FedAvg, Scaffold, and FedROD. The convergence curves are shown in Fig.~\ref{fig:performance-more}. Utilizing a larger local personalization epoch could lead to faster convergence. FedROD may encounter the aggregation performance degradation problem because of longer local training. Compared with other FL algorithms, MAP could obtain better performances under all of these experimental settings.
	
	\begin{table*}
		\renewcommand{\arraystretch}{1.1}
		\centering
		\caption{{\small Performance comparisions under complete classes ($T=150$, $Q=0.2$, averaged across FaMnist, CIFAR-10, and CIFAR-100).}} \label{tab:complete}
		{
			\begin{tabular}{l|c|c|c|c|c|c|c|c|c|c|c}
				\hline
				& FedAvg & FedProx & Scaffold & FedDyn & MOON & pFedMe & Ditto & FedROD & FedRS & FedPHP & MAP \\ \hline
				Aggregation Accuracy & 70.93 & 71.34 & 72.85 & 67.20 & 71.35 & 49.65  & 71.31 & 69.19 & 72.15  & 71.52 & \textbf{73.03} \\ \hline
				Personalization Accuracy & 72.45 & 72.08 & 73.47 & 67.44 & 72.79 & 50.78 & 55.26 & 68.25 & 71.45 & 74.08 & \textbf{75.61} \\
				\hline
			\end{tabular}
		}
	\end{table*}
	
	\subsection{Scenes with Complete Classes}
	One of the limitations of this work is only focusing on the extreme scene of label distribution skew in FL. If we set $\alpha_{c}^k = \frac{m_c^k}{\sum_{j=1}^C m_{j}^k}$ in Eq.~\ref{eq:rs}, where $m_{c}^k$ means the amount of samples in the $c$-th class, we could apply ``RS" to complete classes. We construct scenes with complete classes via decentralizing data by the Dirichlet distribution~\cite{FedPAN,FedDF}, and set the Dirichlet alpha as $0.5$. The used networks and datasets are LeNet on FaMnist, TFCNN on CIFAR-10, and VGG8-BN on CIFAR-100. Other hyper-parameters are kept the same as before. The averaged aggregation and personalization results are listed in Tab.~\ref{tab:complete}. With slight modification, FedRS could also improve the aggregation accuracy, while it is not so effective when compared with Scaffold. The significant personalization improvement of FedPHP shows the pure effectiveness of ``HPM" under complete classes. MAP improves both aggregation and personalization performances, which verifies that combing ``RS" and ``HPM" could also facilitate the learning process under complete scenes. 
	
	\section{Conclusion}
	We study an extreme label distribution Non-I.I.D. scene in FL, i.e., FL with incomplete classes where participating clients could encounter the challenges of missing classes. We point out that the missing classes could harm both the aggregation and personalization performances in FL. On one hand, to tackle the aggregation problem caused by incomplete classes, we propose a novel strategy named ``restricted softmax" to carefully update the proxies of missing classes during local training. On the other hand, to solve the personalization problem caused by incomplete classes, we propose another novel strategy named ``inherited private model" to carefully utilize the historical personalization experience. The two strategies are seamlessly combined as the proposed methods. Our proposed MAP could simultaneously obtain better aggregation and personalization performances. Abundant experimental studies verify the superiorities of our methods. Extending the method to more complex and practical FL scenes may be the future work. Simultaneously considering other factors like fairness in FL is also interesting to study further.

	\section*{Acknowledgements}
	This work is partially supported by the the National Key RD Program of China (Grant No. 2022YFF0712100) and the National Natural Science Foundation of China (Grant No. 61921006, 62006118, 62276131). Thanks to Huawei Noah’s Ark Lab NetMIND Research Team.

	\bibliographystyle{IEEEtran}
	\bibliography{IEEEabrv,map}
	
	%
	
	\begin{IEEEbiography}[{\includegraphics[width=1in,height=1.25in,clip,keepaspectratio]{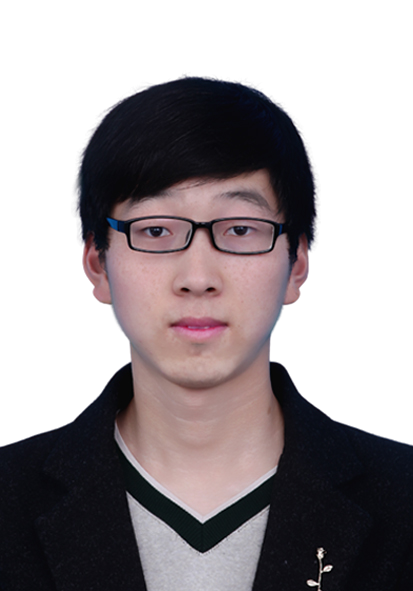}}]{Xin-Chun Li} received his M.Sc. degree with the National Key Lab for Novel Software Technology, the Department of Computer Science and Technology in Nanjing University, China. He is working towards a Ph.D. degree with the School of Artificial Intelligence in Nanjing University, China. Up until now, he has published over 10 papers in national and international journals or conferences such as NeurIPS, CVPR, KDD, ICASSP, SCIS, etc. His research interests lie primarily in machine learning and data mining, including federated learning.
	\end{IEEEbiography}

	\begin{IEEEbiography}[{\includegraphics[width=1in,height=1.25in,clip,keepaspectratio]{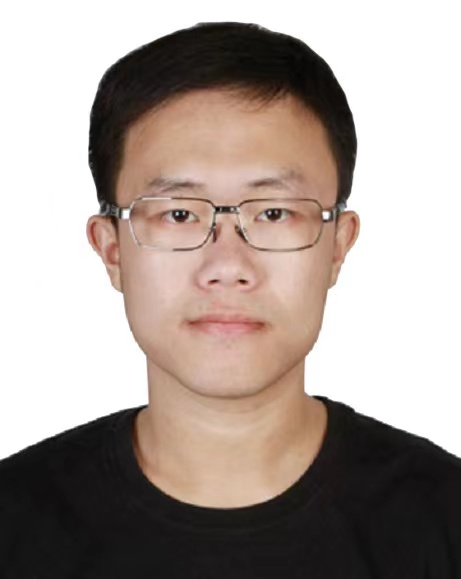}}]{Shaoming Song} received his M.Sc. degree with the Tsinghua National Laboratory for Information Science and Technology, the Department of Automation in Tsinghua University, China. Currently he serves as a senior research engineer at Huawei Noah's Ark Lab. His research interests mainly include machine learning generalization and federated learning.
	\end{IEEEbiography}

	\begin{IEEEbiography}[{\includegraphics[width=1in,height=1.25in,clip,keepaspectratio]{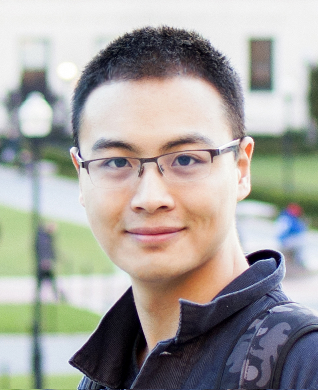}}]{Yinchuan Li} (Member, IEEE) received the B.S. and Ph.D. degrees in electronic engineering from the Beijing Institute of Technology (BIT), Beijing, China, in 2015 and 2020, respectively. From November 2017 to November 2019, he was a Visiting Scholar with the Department of Electrical Engineering, Columbia University, New York, NY, USA. From February 2020 to August 2020, he was a senior technical consultant in Sant\'e Ventures, Austin, TX, USA. He now works at Noah's Ark Lab, Huawei Technologies, Beijing, China, as an AI researcher. His current research interests include machine learning, deep learning, reinforcement learning and sparse signal processing. Dr. Li received the Excellent Paper Award at the 2019 IEEE International Conference on Signal, Information and Data Processing.
	\end{IEEEbiography}

	\begin{IEEEbiography}[{\includegraphics[width=1in,height=1.25in,clip,keepaspectratio]{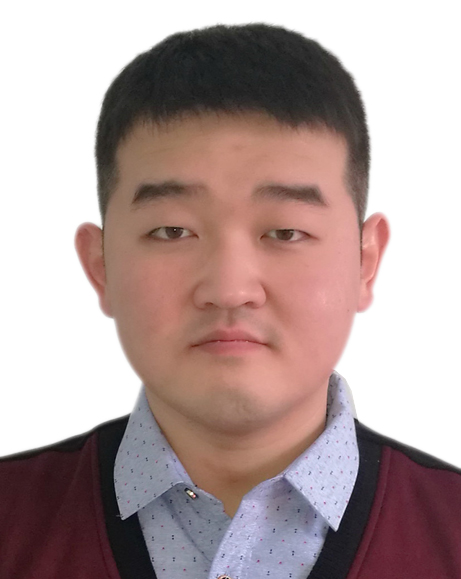}}]{Bingshuai Li} received the B.Sc. and M.Sc. degrees from JiLin University China, in 2014 and 2017, respectively. He is currently an engineer in Huawei Noah’s Ark Lab. His current research interest includes machine learning, federated learning, transfer learning and their applications in Telecommunication network.
	\end{IEEEbiography}
	
	\begin{IEEEbiography}[{\includegraphics[width=1in,height=1.25in,clip,keepaspectratio]{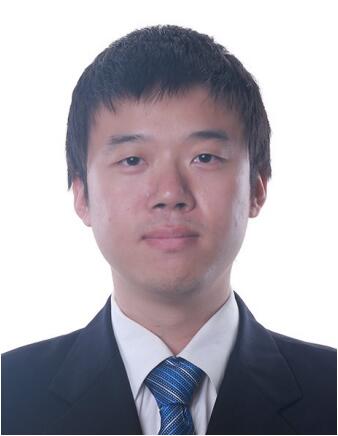}}]{Yunfeng Shao} received the B.S. and Ph.D. degrees in Electronic Engineering from Shanghai Jiao Tong University and University of Chinese Academy of Sciences China, in 2009 and 2014, respectively. He is currently an expert in Huawei Noah’s Ark Lab. His research interests include machine learning with privacy protection, federated learning, transfer learning and their applications in Telecommunication network.
	\end{IEEEbiography}

	\begin{IEEEbiography}[{\includegraphics[width=1in,height=1.25in,clip,keepaspectratio]{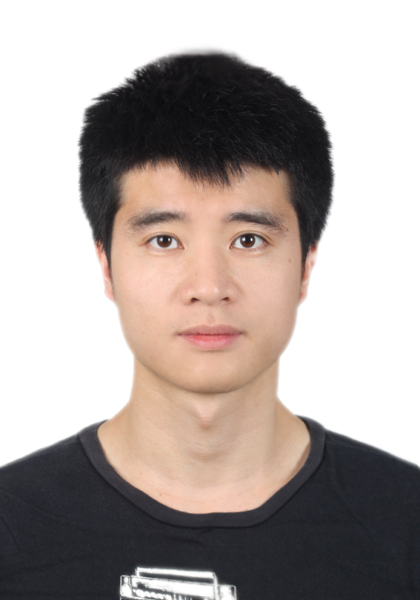}}]{Yang Yang} received the Ph.D. degree in computer science, Nanjing University, China in 2019. At the same year, he became a faculty member at Nanjing University of Science and Technology, China. He is currently a Professor with the school of Computer Science and Engineering. His research interests lie primarily in machine learning and data mining, including heterogeneous learning, model reuse, and incremental mining.  He has published prolifically in refereed journals and conference proceedings, including IEEE Transactions on Knowledge and Data Engineering (TKDE), ACM Transactions on Information Systems (ACM TOIS), ACM Transactions on Knowledge Discovery from Data (TKDD), ACM SIGKDD, ACM SIGIR, WWW, IJCAI, and AAAI. He was the recipient of the the Best Paper Award of ACML-2017. He serves as PC/SPC in leading conferences such as IJCAI, AAAI, ICML, NeurIPS, etc.
	\end{IEEEbiography}

	\begin{IEEEbiography}[{\includegraphics[width=1in,height=1.25in,clip,keepaspectratio]{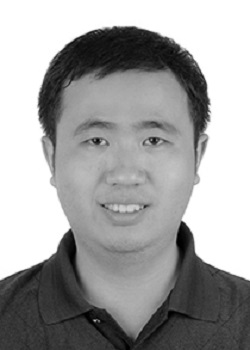}}]{De-Chuan Zhan} joined in the LAMDA Group on 2004 and received his Ph.D. degree in Computer Science from Nanjing University in 2010, and then serviced in the Department of Computer Science and Technology of Nanjing University as an Assistant Professor from 2010, and as an Associate Professor from 2013. Then he joined the School of Artificial Intelligence of Nanjing University as a Professor from 2019. His research interests mainly include machine learning and data mining, especially working on mobile intelligence, distance metric learning, multi-modal learning, etc. Up until now, he has published over 60 papers in national and international journals or conferences such as TPAMI, TKDD, TIFS, TSMSB, IJCAI, ICML, NIPS, AAAI, etc. He served as the deputy director of LAMDA group, NJU, and the director of AI Innovation Institute of AI Valley, Nanjing, China.
	\end{IEEEbiography}
	
	
	

\end{document}